\documentclass{article}

 \usepackage[preprint]{neurips_2026}

\usepackage{amsmath}
\usepackage{amssymb}
\usepackage{mathtools}
\usepackage{amsthm}

\usepackage[utf8]{inputenc} 
\usepackage[T1]{fontenc}    
\usepackage{hyperref}       
\usepackage{url}            
\usepackage{booktabs}       
\usepackage[capitalize,noabbrev]{cleveref}
\usepackage{marvosym}
\crefname{section}{Sec.}{Secs.}
\Crefname{section}{Sec.}{Secs.}

\crefname{figure}{Fig.}{Figs.}
\Crefname{figure}{Fig.}{Figs.}

\crefname{table}{Tab.}{Tabs.}
\Crefname{table}{Tab.}{Tabs.}

\usepackage{makecell}
\usepackage{enumitem}
\usepackage{amsfonts}       
\usepackage{nicefrac}       
\usepackage{microtype}      
\usepackage{xcolor}         
\usepackage[table]{xcolor}
\usepackage{multirow}
\usepackage{pifont}
\newcommand{\cmark}{\ding{51}}
\newcommand{\xmark}{\ding{55}}
\usepackage{graphicx}

\title{Enhancing In-context Panoramic Generation via Geometric-aware Pretraining}

\author{%
\vspace{0.5em}%
Haoran Feng\textsuperscript{1,2}\footnotemark[1] \quad
Ruiyang Zhang\textsuperscript{1,3}\footnotemark[1] \quad
Longyi Zhang\textsuperscript{2} \quad
Dizhe Zhang\textsuperscript{1}\textsuperscript{\Letter}\footnotemark[2] \quad 
Lu Qi\textsuperscript{1,4}\textsuperscript{\Letter} \\
\vspace{0.5em}\small
\textsuperscript{1} Insta360 Research\quad
\textsuperscript{2} Tsinghua University\quad
\textsuperscript{3} Beihang University\quad
\textsuperscript{4} Wuhan University\quad
\\
}

\begin{document}

\footnotetext[0]{%
$^*$ Equal Contribution \hspace{5pt}
$^\dagger$ Project Lead \hspace{5pt}
\textsuperscript{\Letter} Corresponding Author \hspace{5pt}
}

\maketitle
\vspace{-1.5em}
\begin{figure*}[h]
    \centering
    \includegraphics[width=\linewidth]{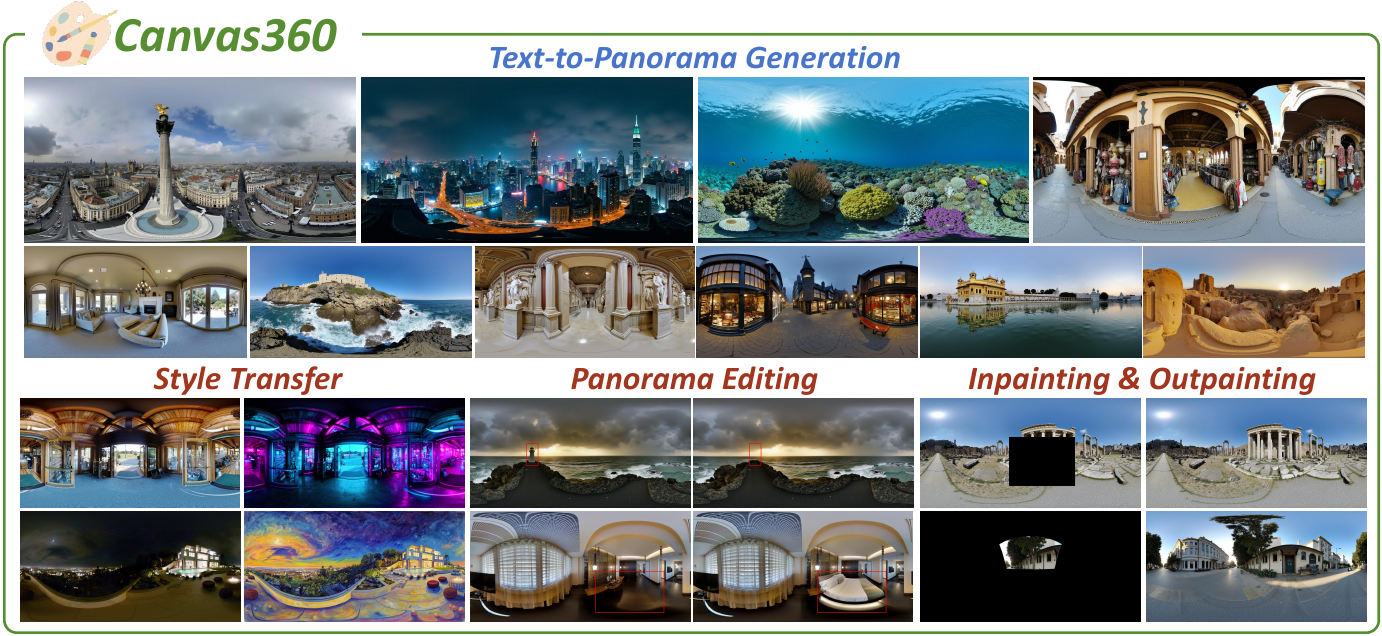}
    \caption{
    Visualization of \textit{Canvas360}'s results. The examples cover text-to-panorama generation, inpainting, outpainting, panorama editing, and style transfer. These results demonstrate that \textit{Canvas360} achieves strong generative performance, captures a rich panoramic prior, and supports a wide range of downstream applications. Additional results are provided in~\cref{appendix:more_app}.
    }
    \label{fig:teaser}
\end{figure*}

\begin{abstract}

In this work, we present \textit{Canvas360}, a two-stage framework for in-context panoramic generation that combines geometry-aware pretraining with downstream task-specific fine-tuning.
To address the lack of large-scale, high-quality training data tailored to in-context panoramic tasks, 
we propose \textit{Canvas360Dataset}, 
a collection of 1M high-quality paired panoramic samples for style transfer, inpainting, outpainting, and editing, enabling effective supervision across diverse in-context generation scenarios.
On the modeling side, \textit{Canvas360} enhances text-to-panorama generation through parallel depth generation, velocity circular padding, and similarity loss regularization, enabling the model to learn geometry-aware representations, capture object distortion details, and improve geometric consistency and global coherence.
Furthermore, empowered by strong panoramic priors, \textit{Canvas360} enables a unified in-context panoramic generation framework that supports diverse downstream tasks via token-level concatenation, surpassing prior methods in both task coverage and modeling flexibility.
Extensive experiments show that \textit{Canvas360} improves panoramic image fidelity, achieving particularly strong performance on the panorama-specific FAED metric and competitive or leading results across the reported quantitative evaluations.
More information can be found on our project page: \url{https://zry000.github.io/Canvas360/}.

\end{abstract}
\section{Introduction}


%

%

With the rapid progress of panoramic text-to-image generation models~\citep{ye2024diffpano, panfusion, worldgen, smgd, ni2025makes, bar2023multidiffusion, li2023panogen, shi2023mvdream, mvdiffusion}, in-context editing has emerged as a natural extension beyond basic text-to-panorama, enabling image generation conditioned jointly on user-provided images and textual prompts~\citep{brooks2023instructpix2pix, flux1kontext, liu2025step1x, suvorov2022resolution}.
This capability underpins a wide range of interactive applications, including content-aware editing~\citep{banana,seededit, wu2025omnigen2} and immersive scene manipulation~\citep{deng2025emerging, yu2025wonderworld}.

Despite these advances, the dominant equirectangular projection (ERP) representation for panoramic images inherently exhibits latitude-dependent distortions, posing challenges for geometry-consistent editing. 
Existing panoramic image editing methods~\citep{yang2025omni, zhong2025se360} attempt to mitigate this issue through distortion-aware designs, such as cube-map-based editing~\citep{yang2025omni} or 3D spherical positional embeddings~\citep{zhong2025se360}. 
Nevertheless, we empirically observe that these approaches still struggle to preserve geometric consistency in the underlying 3D scene structure when operating on ERP panoramas.


Inspired by common practices in perspective visual generation, prior works often introduce depth constraints as explicit geometric priors during training~\citep{huang2025unityvideo, bai2025geovideo, bhat2024loosecontrol, zhang2023jointnet, yu2025wonderworld}.
However, the geometric formulation of depth in panoramic imagery differs from that in planar image settings.
While perspective images define depth along the Cartesian Z-axis, panoramic scenes are naturally represented in spherical space, where depth corresponds to radial distance from the camera center. 
Therefore, a natural question arises: \textit{How can depth priors be formulated under spherical geometry to preserve geometric consistency in in-context panoramic image generation?}

To address this, we propose \textit{Canvas360}, a two-stage in-context panoramic generation framework with geometry-aware pretraining and unified in-context fine-tuning. During pretraining, large-scale RGB panoramas are paired with depth predictions to form RGB–depth data. Latents from both modalities are concatenated and processed by a Flow Transformer, with flow-matching objectives applied to each. Positional offsets and a similarity loss ensure RGB and depth representations remain distinct, while velocity circular padding enforces spherical continuity and boundary consistency. 
In fine-tuning, we train a unified in-context panoramic generation model that jointly supports four tasks: \textbf{style transfer}, \textbf{inpainting}, \textbf{outpainting} and \textbf{editing}. 
Depth is discarded, and the model is trained on high-quality downstream in-context data. Token-level concatenation is adopted to unify heterogeneous contextual conditions, following prior in-context image generation approaches~\citep{flux1kontext, flux1fill}.

%

Moreover, progress in this field has long been hindered by data scarcity. To address this limitation, we curate a high-quality panoramic dataset of 100K indoor and outdoor scenes by building on existing resources~\cite{Matterport3D, feng2025dit360} and leveraging state-of-the-art generation models.
Building on this seed set, we develop a scalable data synthesis pipeline that further produces 900K paired samples for downstream in-context panorama generation tasks—including outpainting (250K), inpainting (250K), style transfer (200K), and panorama editing (200K)—providing a foundation for large-scale model scaling.

To demonstrate the effectiveness of our training pipeline, we conduct extensive experiments on five tasks, including text-to-panorama generation, style transfer, inpainting, outpainting, and editing.
%
%
Experimental results show that Canvas360 improves panorama-specific fidelity and boundary consistency, with leading FAED performance and competitive overall scores on the validation set.
%
%
Our main contributions are summarized as follows:
\begin{itemize}[leftmargin=*, topsep=2pt, itemsep=1pt, parsep=0pt, partopsep=0pt]
    \item We propose \textit{Canvas360}, a two-stage framework that integrates geometry-aware text-to-panorama pretraining with unified downstream in-context fine-tuning. By leveraging large-scale, depth-augmented panoramic data along with curated downstream datasets, \textit{Canvas360} achieves improved spatial consistency and geometric fidelity in in-context panoramic image generation.
    \item We introduce a geometry-aware pretraining strategy based on parallel RGB–depth generation, regularized by a similarity loss between RGB and depth latents. Velocity circular padding further enforces boundary consistency and spherical continuity, benefits that transfer effectively to downstream in-context tasks through fine-tuning on noise-free data.
    \item We design a scalable data synthesis pipeline and propose \textit{Canvas360Dataset}, a 1M-scale dataset for in-context panoramic generation which, to our knowledge, is the most comprehensive to date, spanning four distinct tasks: outpainting, inpainting, style transfer, and panorama editing. Building upon this dataset, we train a unified in-context generation model that jointly learns all four tasks within a single framework, achieving broad task coverage and strong generalization across diverse in-context panoramic scenarios.
    \item Extensive quantitative and qualitative evaluations on both basic text-to-panorama generation and in-context generation tasks demonstrate that \textit{Canvas360} achieves strong performance in boundary consistency, panorama-specific fidelity, and overall perceptual quality.
\end{itemize}

\section{Related Work}
\label{sec:related_work}

\begin{figure*}[t]
    \centering
    \includegraphics[width=\linewidth]{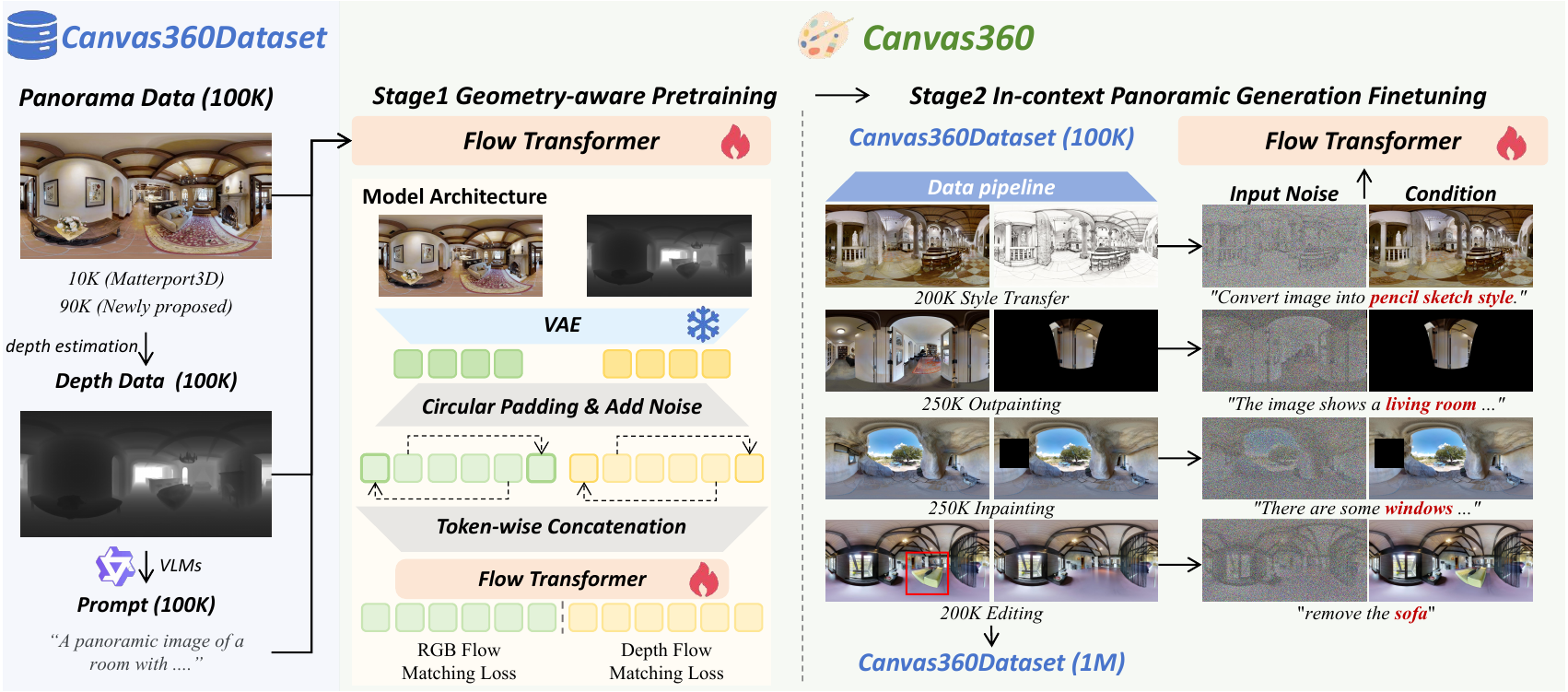}
    \caption{
    Overview of the \textit{Canvas360} two-stage training pipeline. The pipeline is built on \textit{Canvas360Dataset}, which consists of 100K annotated RGB–depth panoramas and 900K in-context generation samples spanning four tasks: style transfer, outpainting, inpainting, and editing. Pretraining is performed on the 100K RGB–depth set using parallel depth generation with velocity circular padding to instill geometric understanding. The unified finetuning stage leverages token-level concatenation to handle diverse contextual conditions and is trained on the 900K downstream samples.
    }
    \label{fig:pipeline}
    \vspace{-0.2in}
\end{figure*}

\noindent\textbf{Large-scale Diffusion and Flow-Matching Models.}
Diffusion models have become the dominant paradigm for image generation~\citep{vae,gan}, achieving high-quality and diverse synthesis by reversing a gradual noising process~\citep{dhariwal2021diffusionbeatgans, glide, imagen, dalle2}. Latent Diffusion Models (LDMs)~\citep{ldm} enable scalable high-resolution synthesis via denoising in a compressed latent space~\citep{sdxl}, while recent transformer-based architectures with explicit positional encodings and global self-attention further improve scalability and performance and are increasingly adopted in large-scale text-to-image systems~\citep{dit, transformer, flux, sd3, repa, sit}. In parallel, flow-matching–based models provide a continuous-time alternative by learning velocity fields that transport noise to data distributions~\citep{lipman2022flow, rectified_flow}, and have been successfully applied in recent large-scale models~\citep{flux, sd3}. 


\noindent\textbf{In-context Panoramic Generation.}
In-context image generation~\citep{flux1kontext, wu2025omnigen2} leverages contextual inputs beyond text, such as reference images, depth maps, masks, or edge cues, and has achieved substantial success on perspective images across tasks including style transfer~\citep{zhang2023inversion}, inpainting~\citep{flux1fill, suvorov2022resolution}, outpainting~\citep{cheng2022inout}, editing~\citep{brooks2023instructpix2pix, liu2025step1x, seededit}, and object manipulation~\citep{deng2025emerging}.

For panoramic images, early efforts rely on multi-view stitching~\citep{fang2023ctrl,hollein2023text2room,yu2023long,bar2023multidiffusion,lee2023syncdiffusion,li2023panogen,shi2023mvdream,mvdiffusion,park2025spherediff,yang2025omni} or cube-map representations~\citep{song2023roomdreamer,ye2024diffpano,huang2025dreamcube,kalischek2025cubediff}, which suffer from view inconsistency and boundary artifacts. 
More recent methods train directly on equirectangular panoramas~\citep{chen2022text2light,shum2023conditional,zhang2023diffcollage,feng2023diffusion360,ai2024dream360,wang2024customizing,yang2024dreamspace,panfusion,worldgen,smgd,hunyuanworld,ni2025makes,par,matrix3d} or introduce spherical-aware convolutions~\citep{smgd,park2025spherediff,panfusion}, but remain constrained by limited data quality. DiT360~\citep{feng2025dit360} addresses these issues via hybrid training on large-scale, high-quality data, enabling sharp details and correct polar distortion.
Existing in-context panoramic methods typically rely on sphere-specific designs, such as cube maps~\citep{yang2025omni} or 3D spherical positional encodings~\citep{zhong2025se360}, and train directly on downstream tasks.
In contrast, we focus on large-scale, high-quality pretraining to learn strong spatial and geometric priors, enabling a unified in-context panoramic generation model that supports diverse downstream tasks within a single framework.

\section{Method}


\subsection{Preliminaries}

\noindent\textbf{Flow Matching.}
Flow Matching (FM)~\cite{lipman2022flow, rectified_flow, geng2025mean} is a continuous-time generative modeling paradigm that has been widely adopted by recent state-of-the-art image generation models~\cite{flux, flux1kontext, sd3} and video generation models~\cite{sora2, veo3, wan2025wan, kong2024hunyuanvideo}.

Let \( x_0 \sim \pi_0 \) denote data from the data distribution, and \( x_1 \sim \pi_1 \) denote noise drawn from a prior distribution (e.g., Gaussian). 
In this paper, we follow the Rectified Flow~\cite{rectified_flow} linear interpolation
\begin{equation}
x_t = (1 - t) x_0 + t x_1, \quad t \in [0,1].
\end{equation}
Flow Matching trains a parameterized model \( v_\theta(x_t, t) \) to regress the velocity field $v=x_1-x_0$ by minimizing a loss function defined as
\begin{equation}
\mathcal{L}(\theta)
= \mathbb{E}_{t,\, x_0 \sim \pi_0,\, x_1 \sim \pi_1}
\left[ \left\| v_\theta(x_t, t) - (x_1-x_0) \right\|^2 \right].
\end{equation}



\noindent\textbf{Diffusion Transformer.}
Flow Transformer architectures used in recent flow-matching–based generative models closely resemble the Diffusion Transformer (DiT)~\citep{dit, feng2025dit360}, inheriting its transformer-based design for modeling continuous-time generative dynamics. DiT adopts a transformer backbone~\citep{vit} to operate on sequences of latent image tokens encoded by a variational autoencoder~\citep{vae}. 
Concretely, an input image is mapped to a token sequence \( X \in \mathbb{R}^{N \times d} \), where \( N \) denotes the sequence length and \( d \) is the embedding dimension. 
To capture spatial structure, DiT employs Rotary Positional Embeddings (RoPE)~\citep{su2024rope}, which inject coordinate-dependent rotations into token representations, enabling parameter-efficient encoding of both relative and absolute positional information. To support multiple image inputs for in-context generation, prior works extend this design with 3D RoPE~\citep{flux1kontext}, indexing latent tokens by spatiotemporal coordinates \((T,H,W)\) to preserve structural alignment across contextual inputs.


\subsection{Geometry-aware Text-to-Panorama Pretraining}
\label{sec: basemodel}

In-context panoramic image generation demands stronger spatial understanding and stricter geometric consistency than standard text-to-panorama synthesis. 
To equip the model with these capabilities, we leverage a large-scale, high-quality, depth-augmented dataset and introduce geometry-aware training strategies that explicitly enforce depth reasoning and panoramic boundary consistency.
More detailed analysis of the geometry-aware training strategies is provided in~\cref{appendix:geometry-aware_training}.


\noindent\textbf{Parallel Depth Generation.}
%
Depth maps provide an explicit geometric representation of 3D scene structure for enhancing spatial understanding, and are more prevalent than other geometric cues in monocular settings~\citep{lin2025depth, tan2026maskeddepthmodelingspatial}. Leveraging depth as auxiliary supervision is therefore a natural and effective choice for improving spatial awareness and geometric fidelity in panoramic image generation~\citep{wu2023panodiffusion}. 
Inspired by prior work~\citep{qi2024unigs, wu2023panodiffusion}, we train the model to generate RGB panoramas and depth maps in parallel, enabling the model to learn geometry-aware panoramic representations under spherical scene structure.

Specifically, we obtain depth maps using DAP~\citep{lin2025depth} and adopt \textbf{sequence concatenation}~\citep{flux1kontext} to combine RGB and depth information by appending post-VAE depth tokens to the RGB token sequence, as demonstrated in ~\cref{fig:pipeline}.
Let $\mathbf{x}_{\text{rgb}} \in \mathbb{R}^{N \times d}$ and $\mathbf{x}_{\text{depth}} \in \mathbb{R}^{N \times d}$ denote the post-VAE token sequences of the RGB image and the depth image, respectively. Sequence concatenation is defined as
\begin{equation}
\mathbf{x} = \left[ \mathbf{x}_{\text{rgb}} \, ; \, \mathbf{x}_{\text{depth}} \right],
\end{equation}
where $[\cdot \, ; \, \cdot]$ denotes concatenation along the token dimension. The flow-matching loss is computed independently for each modality.

To disambiguate RGB and depth latents in positional encoding, we introduce a constant offset along the first dimension of the 3D RoPE embeddings for depth tokens.
Let $\mathbf{u} = (T,H,W)$ denote the positional encoding coordinates. 
We define
\begin{equation}
\mathbf{u_0} = (0,H,W), \ \mathbf{u_1} = (T_d,H,W), \ T_d>0, T_d \in \mathbb{N}, 
\end{equation}
where $\mathbf{u_0}, \mathbf{u_1}$ correspond to RGB and depth tokens, respectively, and $T_d$ controls the offset.

Unlike prior channel-wise designs~\citep{wu2023panodiffusion}, our approach adopts a simpler token-wise formulation with positional offsets to separate RGB and depth. This design enables seamless integration with large pretrained models such as FLUX.1-Kontext~\citep{flux1kontext}, allowing effective reuse of their general visual and generative capabilities.

\begin{figure*}[t]
    \centering
    \includegraphics[width=\linewidth]{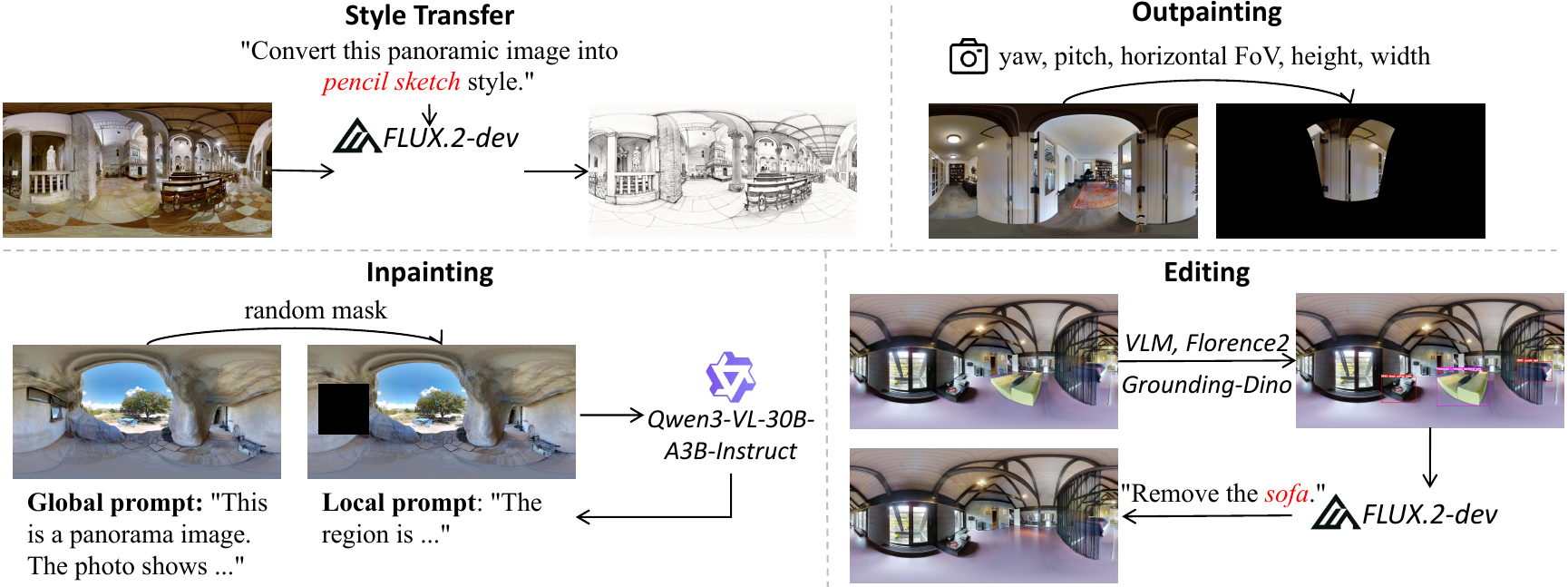}
    \caption{
    Data synthesis pipeline for four in-context panoramic image generation tasks.
    (a) Style Transfer: We directly apply FLUX.2-dev~\citep{flux2} to generate stylized panoramic images across 12 styles.
    (b) Outpainting: We sample random camera parameters to generate diverse perspective-view masks on panoramic images.
    (c) Inpainting: We randomly crop rectangular regions on panoramas and provide two complementary textual annotations: a global prompt describing the entire scene and a local prompt describing the masked region.
    (d) Editing: We employ vision–language models and grounding models to localize objects in panoramas via bounding boxes, and then leverage FLUX.2-dev to remove the targeted objects.
    }
    \label{fig:data_pipeline}
    \vspace{-0.2in}
\end{figure*}

\paragraph{Velocity Circular Padding.}
To address boundary continuity in panoramic image generation, prior
work~\citep{feng2025dit360,zhong2025se360} applies circular padding to
panorama latents.
However, simply copying boundary columns does not explicitly
inform the model that the copied boundary tokens are adjacent on the sphere.
Instead, it mainly increases the optimization weight of boundary regions.
We introduce velocity circular padding to expose this wrap-around adjacency
during velocity prediction. 
Concretely, after reshaping the interpolated latent
$x_t \in \mathbb{R}^{N\times d}$ into $x_t \in \mathbb{R}^{H\times W\times d}$,
we index the original longitude columns as $1,\ldots,W$ and append two ghost
columns with longitude indices $0$ and $W{+}1$. 
Before the transformer
computation, we synchronize the ghost-column features with their circular
counterparts:
\[
\tilde{x}_t^{0} = x_t^{W}, \qquad
\tilde{x}_t^{j} = x_t^{j},\; j=1,\ldots,W, \qquad
\tilde{x}_t^{W+1} = x_t^{1}.
\]
The target velocity is synchronized in the same way:
\[
\tilde{v}^{0} = v^{W}, \qquad
\tilde{v}^{j} = v^{j},\; j=1,\ldots,W, \qquad
\tilde{v}^{W+1} = v^{1}.
\]
The padded sequence therefore uses feature synchronization across the
$0^\circ/360^\circ$ boundary while preserving continuous longitude indices
$0,1,\ldots,W,W{+}1$. 
In this way, the model observes the local adjacency
between columns $(0,1)$ and $(W,W{+}1)$, while the synchronized features impose
the spherical equivalences $0 \equiv W$ and $W{+}1 \equiv 1$. 
We compute the
flow-matching loss on the padded velocity field to explicitly supervise
boundary-consistent velocity prediction.

\noindent\textbf{Similarity Loss Regularization.}
During training, we observe that the model can converge to a degenerate local optimum in which the predicted RGB and depth outputs become overly similar. To explicitly encourage modality-specific representations, we introduce a \textbf{similarity loss} as a regularization term that penalizes excessive correlation between the RGB and depth predictions.

The similarity loss is defined as the squared correlation between the predicted velocity fields of the two modalities.
Let $\mathbf{v}_{\text{rgb}}$ and $\mathbf{v}_{\text{depth}}$ denote the predicted velocity fields for the RGB and depth branches, respectively. We formulate the loss as
\begin{equation}
\mathcal{L}_{\text{sim}} =\mathbb{E} \left[
\left(
\frac{
\langle \mathbf{v}_{\text{rgb}}, \mathbf{v}_{\text{depth}} \rangle
}{
\|\mathbf{v}_{\text{rgb}}\|_2 \, \|\mathbf{v}_{\text{depth}}\|_2
}
\right)^2
\right],
\end{equation}
where $\langle \cdot, \cdot \rangle$ denotes the inner product.
%
The overall training objective is given by
\begin{equation}
\mathcal{L}
=
\mathcal{L}_{\text{FM}}
+
\lambda \, \mathcal{L}_{\text{sim}},
\end{equation}
where $\mathcal{L}_{\text{FM}}$ denotes the flow matching loss and $\lambda$ controls the strength of the similarity regularization.



\subsection{Unified In-context Panoramic Generation Finetuning}



The pretraining stage equips the model with geometry-aware spatial priors using large-scale depth-augmented panoramic data.
Building upon this pretrained text-to-panorama model, we fine-tune a unified in-context panoramic generation model across four representative tasks: \textbf{style transfer}, \textbf{inpainting}, \textbf{outpainting}, and \textbf{editing}.
During fine-tuning, we remove depth inputs and reformulate the model to rely solely on RGB-based in-context conditions, thereby forcing it to operate under appearance-only supervision while implicitly inheriting the spatial structure learned during pretraining. 
As shown in~\cref{fig:pipeline}, this design enables adaptation to downstream tasks without requiring explicit geometric supervision.
Compared to prior works that train task-specific models or directly learn from limited task data~\citep{yang2025omni, zhong2025se360}, our approach leverages pretrained spatial priors to enable a unified and more versatile generation framework.
We formulate all tasks under a shared training tuple:
\begin{equation}
\mathcal{D} = (\mathbf{x}_{\text{cxt}}, c, \mathbf{x}_{\text{tgt}}),
\end{equation}

where $\mathbf{x}_{\text{cxt}}$ and $\mathbf{x}_{\text{tgt}}$ denote the post-VAE latents of the context and target panorama images respectively, and $c$ represents the text prompt. For style transfer and editing, $\mathbf{x}_{\text{cxt}}$ corresponds to the full input panorama. For inpainting and outpainting, $\mathbf{x}_{\text{cxt}}$ is a masked panorama latent with missing regions removed. This unified representation enables all four tasks to be handled within a single framework.

Based on this formulation, we adopt the same design principles as in the pretraining stage~(\cref{sec: basemodel}), including sequence concatenation with positional offset. Specifically, the input token sequence is constructed as:
\begin{equation}
\mathbf{x} = \left[ \mathbf{x}_{\text{tgt}} \, ; \, \mathbf{x}_{\text{cxt}} \right],
\end{equation}
\begin{equation}
\mathbf{u}_{\text{tgt}} = (0,H,W), \ \mathbf{u}_{\text{cxt}} = (T_c,H,W), \ T_c>0, T_c \in \mathbb{N}, 
\end{equation}
where $\mathbf{u}_{\text{tgt}}$ and $\mathbf{u}_{\text{cxt}}$ denote the positional coordinates of the target and context tokens, respectively. The offset $T_c$ separates the two token groups in positional space, enabling the model to differentiate their semantic roles while maintaining spatial correspondence.

\subsection{Data Pipeline}
\label{Sec: data}


In this section, we present the \textit{Canvas360Dataset} synthesis pipeline, a 1M-sample dataset designed for in-context panoramic generation. As shown in ~\cref{fig:pipeline}, the dataset builds upon a 100K-sample pilot set, sourced from Matterport3D, the web, and state-of-the-art panoramic generation models. Captions describing the full panorama are generated for the pilot set using vision-language models (VLMs). From this pilot set, we curate 900K downstream in-context samples across four tasks: style transfer, outpainting, inpainting, and editing.

\noindent\textbf{Style Transfer Data.}
To curate style transfer data, we use FLUX.2-dev~\citep{flux2} with style-specific prompts to synthesize stylized panoramas for each input panorama, as shown in~\cref{fig:data_pipeline}.
Since style transfer mainly alters pixel-level appearance without requiring global spatial reasoning, FLUX.2-dev is well-suited for large-scale curation.
Using this pipeline, we generate 200K style transfer samples.

\noindent\textbf{Outpainting Data.}
Outpainting samples are generated by sampling diverse perspective views from each panorama and deriving inverse-projection masks. 
For each sample, camera parameters, including yaw, pitch, field of view, height, and width, are randomly drawn from predefined priors, as shown in~\cref{fig:data_pipeline}. 
After applying the sampled yaw shift, we project a centered perspective view and inversely project it back to the panorama to obtain the visible-region mask. 
Each sample contains the yaw-shifted panorama, the mask, and a textual caption. 
This pipeline produces 250K outpainting samples, enabling perspective-to-panorama generation across diverse camera settings.

\noindent\textbf{Inpainting Data.}
Following prior work~\citep{ldm, suvorov2022resolution}, we consider two inpainting settings: global-prompted and local-prompted. 
The former uses panorama-level prompts to guide masked-region reconstruction, while the latter uses prompts describing only the masked content. 
For both settings, we generate rectangular masks with random area ratios, aspect ratios, and locations, as shown in~\cref{fig:data_pipeline}, with larger masks for global-prompted samples to capture broader contextual dependencies. 
Global-prompted samples use panorama-level captions, whereas local-prompted samples use Qwen3-VL-30B-A3B-Instruct~\citep{bai2025qwen3vltechnicalreport} to generate captions for the masked regions. 
This procedure yields 250K inpainting samples.

\noindent\textbf{Editing Data.}
To synthesize editing data, we follow the SE360 pipeline~\citep{zhong2025se360}, which combines VLM captioning and fused grounding with Florence2~\citep{xiao2024florence} and GroundingDino~\citep{liu2024grounding}. 
We use it to ground objects in panoramas and obtain both bounding-box and segmentation masks.
Based on these annotations, we use the bounding boxes and grounding masks to guide object erasure, and adopt FLUX.2~\citep{flux2} to remove the targeted objects from panoramic images.
For challenging cases with small objects, fine structures, or complex backgrounds, we further use NanoBanana~\citep{banana} for refinement to better preserve local details and background consistency.
We invert original-edited pairs to obtain both erasure and addition samples, addressing the lack of panoramic training in existing models and promoting geometry-consistent object generation. 
This process yields 200K editing samples, with details in~\cref{appendix:dataset}.
\begin{figure*}[t]
    \centering
    \includegraphics[width=\linewidth]{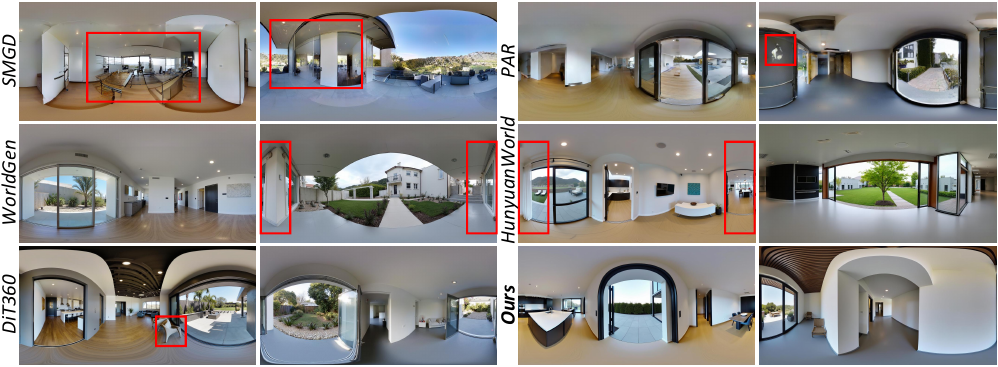}
    \caption{
    Qualitative comparisons for panorama generation, with representative artifacts highlighted in red boxes.
    More results are provided in~\cref{appendix:full_comparison}.
    }
    \label{fig:comparison}
\vspace{-0.2in}
\end{figure*}

\section{Experiments}

\subsection{Setup}
\label{sec:setup}

\textit{Canvas360} is built on FLUX.1-dev~\citep{flux} and fine-tuned via LoRA~\citep{lora}.
For in-context panoramic generation tasks, we train and evaluate on our constructed \textit{Canvas360Dataset}. 
For text-to-panorama generation, to ensure fair comparison, we follow prior work~\cite{feng2025dit360} and use the Matterport dataset~\cite{Matterport3D} for training and validation.
To assess the effectiveness of our approach, we adopt a diverse set of complementary metrics covering realism, diversity, text–image alignment, and perceptual quality, ensuring a comprehensive assessment of model performance. 
More detailed descriptions of the implementation, dataset, and metric definitions are in Appendix~\ref{appendix:experiment_settings}.

\begin{table*}[t]
\centering
\caption{Quantitative comparison results on text-to-panorama generation. Best results are in red and second best are in orange.}
\label{tab:metrics}
\small
\setlength{\tabcolsep}{4pt}
\renewcommand{\arraystretch}{1.05}
\begin{tabular}{lcccccccccc}
\toprule
Methods & 
FID$\downarrow$ &  
\makecell{FID\\\textsubscript{pole}$\downarrow$} & 
\makecell{FID\\\textsubscript{equ}$\downarrow$} &
FAED$\downarrow$ & 
IS$\uparrow$ & 
CS$\uparrow$ & 
\makecell{QA\\\textsubscript{quality}$\uparrow$} &
\makecell{QA\\\textsubscript{aesthetic}$\uparrow$} &
BRISQUE$\downarrow$ &
NIQE$\downarrow$ \\
\midrule
PanFusion       & 124.87 & 182.09 & 108.12 & 11.06 & 1.30 & 28.35 & 3.83 & 3.56 & 27.38 & 4.31 \\
SMGD            & 46.72  & 65.69  & 34.84  & 3.29  & 1.40 & 31.14 & 4.05 & 3.77 & 30.35 & 4.75 \\
PAR             & 47.72  & 76.93  & 27.39  & 2.97  & 1.34 & 33.85 & 3.91 & 3.54 & 32.26 & 4.38 \\
WorldGen        & 67.11  & 79.32  & 33.45  & 3.29  & 1.40 & 34.61 & 4.30 & 3.59 & 32.31 & 4.82 \\
LayerPano3D     & 62.82  & 80.37  & 38.67  & 2.98  & 1.50 & 34.40 & \cellcolor{red!25}4.73 & 3.93 & 33.91 & 3.79 \\
HunyuanWorld    & 76.75  & 106.58 & 41.75  & \cellcolor{orange!25}2.91 & 1.53 & \cellcolor{red!25}34.73 & 4.67 & 3.85 & 39.12 & 5.18 \\
DiT360          & \cellcolor{red!25}42.88 & \cellcolor{red!25}50.88 & \cellcolor{red!25}24.77 & \cellcolor{orange!25}2.91 & \cellcolor{orange!25}1.60 & \cellcolor{orange!25}34.68 & 4.69 & \cellcolor{orange!25}4.19 & \cellcolor{red!25}10.25 & \cellcolor{orange!25}3.72 \\
\midrule
\textbf{Ours}   & \cellcolor{orange!25}44.17 & \cellcolor{orange!25}51.02 & \cellcolor{orange!25}25.96 & \cellcolor{red!25}2.33 & \cellcolor{red!25}1.76 & 34.62 & \cellcolor{orange!25}4.71 & \cellcolor{red!25}4.20 & \cellcolor{orange!25}17.12 & \cellcolor{red!25}3.70 \\
\bottomrule
\end{tabular}
\vspace{-8pt}
\end{table*}

\subsection{Main Results}
\label{sec:main_results}

\noindent\textbf{Qualitative Comparisons.}
We provide qualitative comparisons in \cref{fig:comparison} and highlight artifacts with red boxes. 
SMGD~\citep{smgd} and PAR~\citep{par} explore alternative paradigms based on structural modifications and autoregressive generation, 
but often sacrifice fine-detail fidelity, resulting in cluttered or imprecise outputs.
Works such as WorldGen~\cite{worldgen} and HunyuanWorld~\cite{hunyuanworld} adopt Diffusion Transformers~\cite{dit} as the backbone and achieve substantial improvements, yet still fall short in fine-grained detail for panoramic imagery.
DiT360~\cite{feng2025dit360} further improves fine-detail accuracy with designs such as cube loss, but the cube-to-panorama conversion remains lossy, especially in latent space, leading to residual artifacts and reduced consistency.
In contrast, our method introduces depth to learn geometry-aware panoramic details globally, improving geometric fidelity and producing more accurate renderings that better respect panoramic projection distortions.
More results are provided in Appendix~\ref{appendix:full_comparison}.

\noindent\textbf{Quantitative Comparisons.}
We conduct quantitative evaluations to assess our approach, with results summarized in \cref{tab:metrics}.
\textit{Canvas360} achieves the best FAED score, substantially improving the panorama-specific fidelity metric over prior methods. 
It also obtains the best IS, QA aesthetic, and NIQE scores, and remains competitive across the remaining metrics, ranking second on FID, FID\textsubscript{pole}, FID\textsubscript{equ}, QA quality, and BRISQUE. These results suggest a favorable balance between panorama-aware fidelity, perceptual quality, and geometric consistency.
More results are provided in Appendix~\ref{appendix:full_comparison}.

\begin{figure*}[t]
    \centering
    \includegraphics[width=\linewidth]{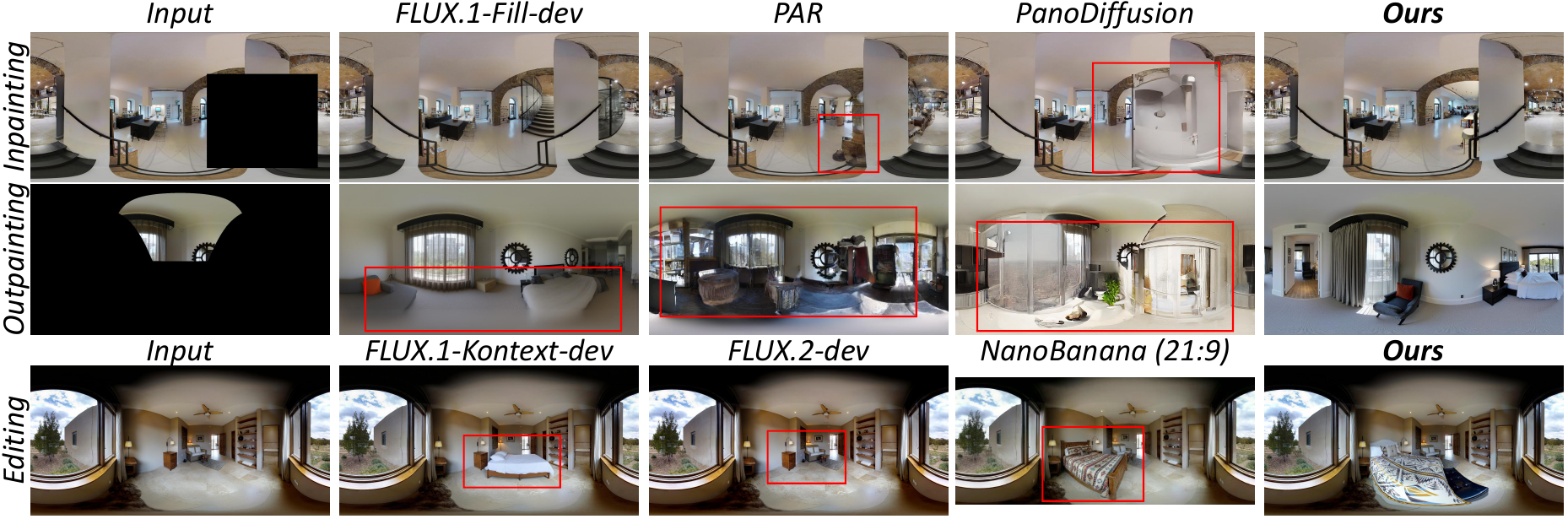}
    \caption{Qualitative comparisons for in-context panoramic generation, with representative artifacts highlighted in red boxes. More results are provided in~\cref{appendix:full_comparison_on_in-context_panoramic_generation}.}
    \label{fig:applications}
    \vspace{-0.2in}
\end{figure*}

\noindent\textbf{In-context Panoramic Generation.}
We present qualitative comparisons on downstream applications in~\cref{fig:applications}, including inpainting, outpainting, and editing. 
For inpainting and outpainting, we compare panorama-specific methods~\cite{par,wu2023panodiffusion} as well as FLUX.1-Fill-dev~\cite{flux1fill}.
The panorama-specific baselines tend to introduce noticeable blur and artifacts, 
whereas the latter is adequate for inpainting with small missing regions, but produces extensive blur in outpainting.
In contrast, \textit{Canvas360} generates panorama-consistent outputs with clean and artifact-free visuals.
For editing, we compare against mainstream editing baselines~\cite{flux1kontext,flux2,banana}. 
While existing methods exhibit some editing capability, they fail to apply the correct panoramic distortion, leading to geometry-inconsistent edits. 
In contrast, \textit{Canvas360} applies correct panoramic distortion to the added content, indicating geometry-aware panoramic priors and validating our method and data pipeline.
More comparisons on downstream tasks can be found in Appendix~\ref{appendix:full_comparison_on_in-context_panoramic_generation}.

\begin{table*}[t]
\centering

\begin{minipage}[t]{0.46\textwidth}
\centering
\caption{User study results on text-to-panorama generation.}
\label{tab:user_study}
\small
\setlength{\tabcolsep}{3pt}
\begin{tabular}{@{}lcccc@{}} 
\toprule
Methods & 
TA $\uparrow$ & 
BC $\uparrow$ & 
PA $\uparrow$ & 
OQ $\uparrow$ \\
\midrule
Matrix-3D     & 23.0\% & 20.0\% & 16.8\% & 15.3\% \\
HunyuanWorld  & \cellcolor{red!25}26.9\% & 20.9\% & 20.5\% & 18.1\% \\
DiT360        & 23.9\% & 28.0\% & 28.6\% & 30.7\% \\
Ours          & 26.2\% & \cellcolor{red!25}31.1\% & \cellcolor{red!25}34.1\% & \cellcolor{red!25}35.9\% \\
\bottomrule
\end{tabular}
\end{minipage}
\hfill
\begin{minipage}[t]{0.50\textwidth}
\centering
\caption{Quantitative ablations for parallel depth generation.}
\label{tab:ablation}
\small
\setlength{\tabcolsep}{3pt}
\begin{tabular}{@{}lcccc@{}} 
\toprule
Methods & 
FID $\downarrow$ &  
FAED $\downarrow$ & 
BRISQUE $\downarrow$ &
NIQE $\downarrow$ \\
\midrule
baseline                    
& \cellcolor{red!25}51.16 & 5.37 & \cellcolor{red!25}14.62 & 3.86 \\
+ depth img                 
& 51.57 & \cellcolor{red!25}4.74 & 17.84 & 3.85 \\
+ pos offset                
& 57.04 & 4.81 & 23.64 & 4.12 \\
+ $\mathcal{L}_{\text{sim}}$    
& 51.48 & 4.93 & 16.23 & \cellcolor{red!25}3.84 \\  
\bottomrule
\end{tabular}
\end{minipage}
\vspace{-0.2in}
\end{table*}


\noindent\textbf{User study.}
To better assess human preference, we conducted a user study comparing our method with several representative baselines~\citep{matrix3d,hunyuanworld,feng2025dit360}.  
We evaluated four criteria: text alignment (TA), boundary continuity (BC), panorama awareness (PA), and overall quality (OQ). %
In total, 71 participants selected their preferred results among different methods on a test set of 10 images.  
As reported in \cref{tab:user_study}, \textit{Canvas360} achieves the highest preference across BC, PA, and OQ, 
demonstrating superior panorama-consistent generation with seamless seam alignment and validating the effectiveness of our method.
Additional details are provided in~\cref{appendix:user_study}.

\subsection{Ablations}
\label{sec:ablations}

We conduct extensive ablation studies to validate the key components of our framework, including parallel depth generation, velocity circular padding, and the \textit{Canvas360} backbone.
More detailed experimental results are provided in~\cref{appendix:ablation}.


\noindent\textbf{Parallel Depth Generation.}
As shown in~\cref{fig:ablations_for_pdg}, starting from a fine-tuned FLUX.1-dev~\cite{flux} baseline, we progressively add depth conditioning, position offsets, and $\mathcal{L}_{\text{sim}}$.
Without these components, the model suffers from geometric distortions and poor panoramic consistency.
Spherical depth improves geometry-aware panoramic modeling, but RGB--depth joint generation can be unstable in certain settings, resulting in over-darkened outputs and visible artifacts.
The position offset mitigates this issue by separating RGB and depth tokens in positional space, and $\mathcal{L}_{\text{sim}}$ further prevents excessive coupling between the two predicted modalities.
Although these stabilization terms are not designed to monotonically improve every individual metric, they reduce degenerate dark-output cases and produce cleaner, more stable panoramic generations.
The quantitative results in~\cref{tab:ablation}, together with the qualitative comparisons in~\cref{fig:ablations_for_pdg}, indicate that the full design improves training robustness, visual stability, and panoramic consistency.

\noindent\textbf{Velocity Circular Padding.}
\cref{fig:ablations_for_vcp} evaluates the effect of velocity circular padding. 
For clearer visualization, we yaw-rotate the inputs by $180^\circ$ to expose the panorama boundary. 
Compared with standard circular padding, our velocity circular padding introduces additional supervision for boundary regions, resulting in better boundary continuity and more accurate edge alignment. 
This validates its importance for maintaining seamless panoramic generation.

\noindent\textbf{Backbone Design.}
~\cref{fig:ablation_for_baseline} compares FLUX.1-dev~\cite{flux} and \textit{Canvas360} fine-tuned under the same setting.
FLUX.1-dev produces blurred, artifact-prone inpainting results and outpainting outputs that are often inconsistent with the conditioning signal and biased toward perspective-image priors.
In contrast, \textit{Canvas360} generates more coherent and panorama-consistent completions, demonstrating that its depth-aware backbone provides stronger panoramic priors and geometry-aware generation capability.





\section{Conclusion}
\label{sec:conclusion}
We presented \textit{Canvas360}, a two-stage in-context framework for panoramic image generation that injects geometry-aware priors through parallel RGB--depth pretraining and transfers them to downstream tasks via unified in-context fine-tuning.
Specifically, we pair large-scale panoramas with predicted depth, fuse RGB and depth latents at the token level, and train a Flow Transformer with positional offsets, similarity regularization, and velocity circular padding to enforce spherical continuity and improve seam alignment.
To alleviate the data bottleneck, we develop a scalable data pipeline and release \textit{Canvas360Dataset}, a 1M-scale dataset covering inpainting, outpainting, style transfer, and panorama editing.
Experiments demonstrate consistent improvements in geometric adherence, seam consistency, and visual fidelity over prior methods, establishing a strong foundation for future scaling and broader panoramic generation applications.

\bibliographystyle{plainnat}
\bibliography{neurips_2026}

\newpage
\section*{Appendix}
\appendix

\section{Analysis of the Geometry-aware Training Strategies}
\label{appendix:geometry-aware_training}

\begin{figure*}[h]
    \centering
    \includegraphics[width=\linewidth]{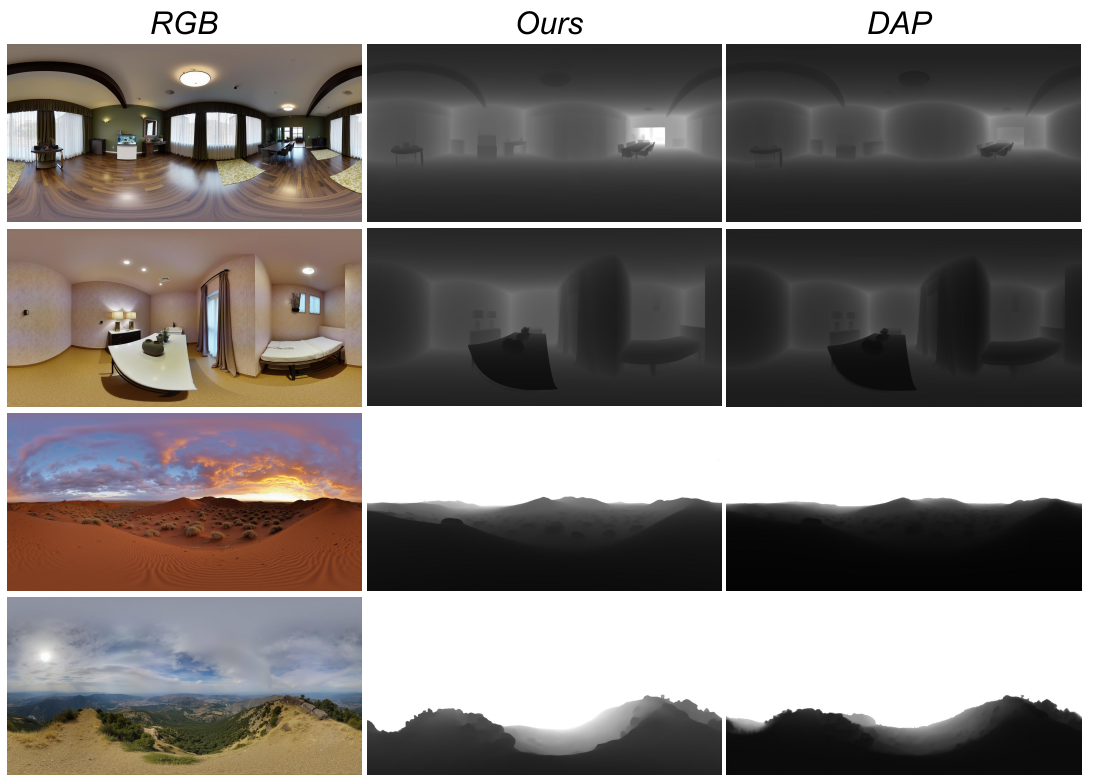}
    \caption{
    Visualization of the predicted depth maps generated by our model. 
    The predicted depth maps are structurally aligned with the corresponding panoramic scenes, indicating that the depth branch provides meaningful geometric guidance during in-context panoramic generation.
    }
    \label{fig:appendix_predicted_depth}
\end{figure*}

\begin{figure*}[h]
    \centering
    \includegraphics[width=\linewidth]{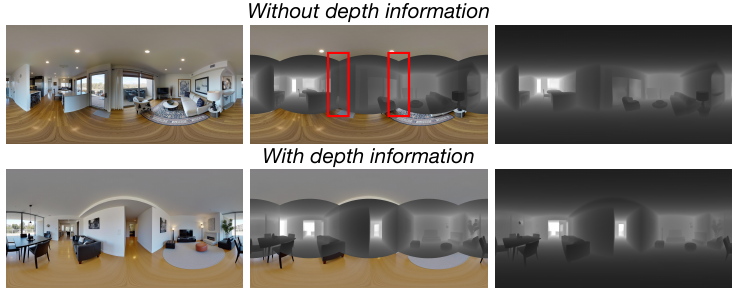}
    \caption{
    Analysis of the geometry prior retained after depth-supervised training. 
    We compare models trained with and without depth supervision. 
    For each generated panorama, we convert it into cubemap faces, retain the four side faces, estimate their depth maps, and stitch them back into the ERP format. 
    The model trained with depth supervision produces more geometrically consistent results, especially around stitching seams and boundary regions.
    }
    \label{fig:appendix_geometry_prior}
\end{figure*}

We further analyze the effect of our geometry-aware training strategies from two complementary perspectives. 
First, we examine whether the model can predict meaningful depth maps during generation. 
Second, we evaluate whether depth-supervised training improves the geometric structure of the generated RGB panoramas themselves.

\noindent\textbf{Visualization of Predicted Depth.}
Our framework jointly generates panoramic RGB images and their corresponding depth maps. 
As shown in~\cref{fig:appendix_predicted_depth}, the predicted depth maps preserve the major scene layout and object-level geometric structures, and are well aligned with the generated panoramic images. 
This demonstrates that the depth branch does not merely produce auxiliary outputs, but captures meaningful geometric information that can serve as effective guidance for panoramic generation.

\noindent\textbf{Geometry Prior Retained After Depth-supervised Training.}
To further examine whether depth supervision improves the geometry of the generated RGB panoramas, we compare a model trained with depth supervision against a counterpart trained without it. 
For the images generated by both models, we convert each panorama into cubemap faces, retain the four side faces while discarding the top and bottom faces, estimate their depth maps, and then stitch the estimated depths back into the original ERP format. 
The results are shown in~\cref{fig:appendix_geometry_prior}. 
The model trained without depth supervision exhibits clear depth inconsistencies, especially near stitching seams and boundary regions. 
In contrast, the model trained with depth supervision produces more coherent depth structures and better cross-view consistency. 
These results indicate that depth-supervised training effectively injects a geometry prior into the model, improving not only the predicted depth maps but also the structural consistency of the generated panoramic images.

\section{More Details on Dataset Construction}
\label{appendix:dataset}

\paragraph{Comparison with Existing Data Pipelines.}
Existing datasets related to panoramic scenes are mainly designed for perception, embodied reasoning, multi-modal understanding, or camera-controlled generation, rather than in-context panoramic generation. 
For example, H*Bench~\citep{yu2025thinking} focuses on embodied visual search with human-annotated reasoning, but does not provide generation-oriented construction or in-context training pairs. 
Puffin-4M~\citep{liao2025puffin} supports camera-controlled generation with explicit camera modeling, yet it does not construct task-driven generative pairs such as inpainting, outpainting, or editing for panoramic in-context learning. 
The 360+x Dataset~\citep{chen2024360+} emphasizes multi-modal data collection and alignment, but lacks task-level generative design and explicit geometry-aware supervision for generation. 
In contrast, our \textit{Canvas360Dataset} is explicitly built for in-context panoramic generation. 
It contains 1M task-driven samples across style transfer, outpainting, inpainting, and editing, with paired input--output data and geometry-aware supervision.

\begin{table*}[ht]
\centering
\caption{
Comparison between \textit{Canvas360Dataset} and existing dataset construction pipelines. 
\textit{Canvas360Dataset} is explicitly designed for in-context panoramic generation, covering multiple generation tasks with paired training data and geometry-aware supervision.
}
\label{tab:appendix_dataset_comparison}
\begingroup
\scriptsize
\setlength{\tabcolsep}{3pt}
\renewcommand{\arraystretch}{1.15}
\begin{tabular}{lccccccc}
\toprule
Data pipelines 
& \shortstack{Generation-\\oriented}
& \shortstack{Panoramic\\Data}
& \shortstack{Explicit Camera\\Modeling}
& \shortstack{Multi-task\\Design}
& \shortstack{Human-\\in-the-loop}
& \shortstack{Language\\Annotation}
& \shortstack{In-context\\Design} \\
\midrule
H*Bench 
& \xmark & \cmark & \cmark & \xmark & \cmark & \cmark & \xmark \\
Puffin-4M 
& \xmark & \xmark & \cmark & \cmark & \xmark & \cmark & \cmark \\
360+x Dataset 
& \xmark & \cmark & \xmark & \cmark & \cmark & \xmark & \xmark \\
Ours 
& \cmark & \cmark & \cmark & \cmark & \cmark & \cmark & \cmark \\
\bottomrule
\end{tabular}
\endgroup
\end{table*}

\paragraph{Quality Control for AI-generated Samples.}
Since current image generation and editing models are not explicitly trained on panoramic data, directly applying them to ERP panoramas may introduce geometry-inconsistent artifacts. 
We mitigate this issue during data construction with task-specific strategies. 
For inpainting and outpainting, all training pairs are derived from real panoramic images, ensuring that the target outputs preserve realistic panoramic geometry. 
For editing, planar image editing models are less reliable for object addition in panoramic scenes. 
Therefore, we use grounding annotations to guide object removal and construct original--edited pairs accordingly; the pairs are then inverted to obtain both erasure and addition samples. 
For style transfer, we compare extracted line drawings between the original and stylized images and remove samples with large structural inconsistencies.

We further conduct manual inspection on a 50K subset of the constructed data, where 48K samples are identified as clean and valid. 
This suggests that the remaining noise is limited and provides an acceptable trade-off for large-scale dataset construction. 
To further examine the impact of such noise, we compare a model fine-tuned on 20K manually cleaned samples with one trained on 20K randomly selected image-editing samples. 
As shown in~\cref{tab:appendix_dataset_cleaning_ablation}, the two models achieve similar performance, indicating that the residual data noise has limited influence on the final model performance.

\begin{table}[ht]
\centering
\caption{
Effect of data cleaning on image-editing performance. 
We compare models trained on 20K randomly selected samples and 20K cleaned samples. 
Lower LPIPS and FAED indicate better performance, while higher PSNR indicates better performance.
}
\label{tab:appendix_dataset_cleaning_ablation}
\begin{tabular}{lccc}
\toprule
Method & LPIPS$\downarrow$ & FAED$\downarrow$ & PSNR$\uparrow$ \\
\midrule
Randomly selected samples & 0.096 & 0.396 & 25.34 \\
Cleaned samples           & \textbf{0.093} & \textbf{0.391} & \textbf{25.70} \\
\bottomrule
\end{tabular}
\end{table}

\paragraph{Pseudo-depth Processing.}
We generate pseudo-depth maps using DAP~\citep{lin2025depth}, a state-of-the-art model for panoramic metric-depth estimation. 
Since DAP predicts absolute depth, extremely large depth values from distant regions may dominate the depth distribution and destabilize training. 
To reduce this effect, we truncate overly large values before normalization. 
Specifically, depth values are clipped at 100 for outdoor scenes and 10 for indoor scenes. 
After truncation, we compare the processed pseudo-depth maps with available ground-truth depth and find that the discrepancy remains small, suggesting that the estimation error is acceptable for our training pipeline. 
Finally, we normalize the depth maps to a fixed range before training, which further mitigates the effect of residual estimation noise and stabilizes RGB--depth co-training.

\section{Experiment Settings}
\label{appendix:experiment_settings}

\begin{figure*}[h!]
    \centering
    \includegraphics[width=1\linewidth]{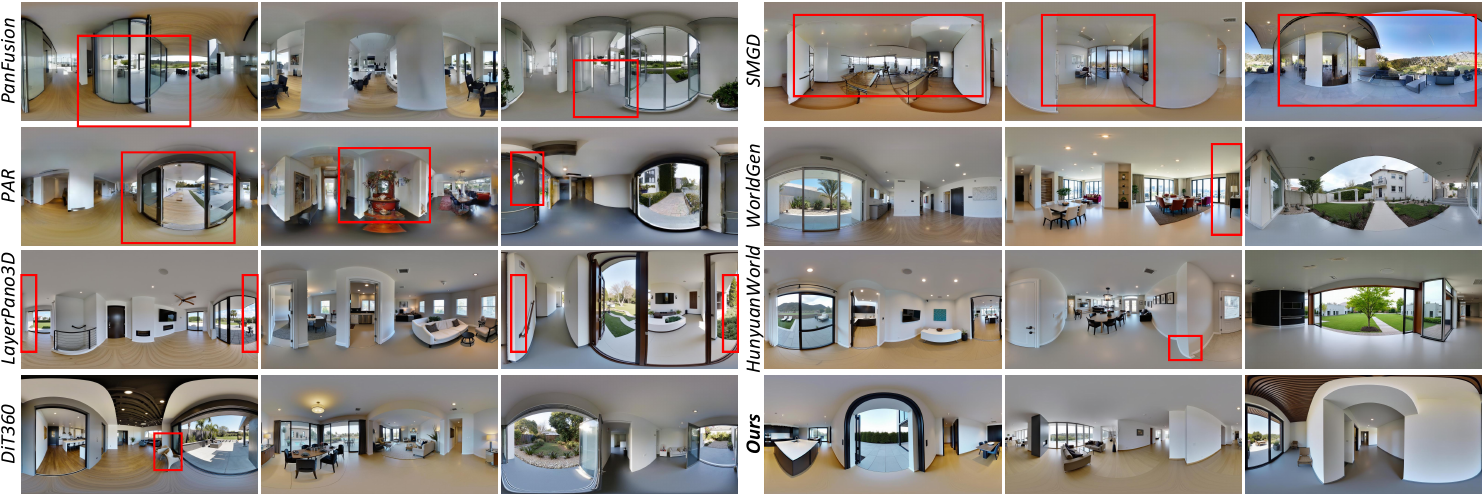}
    \caption{
    Full qualitative comparisons for panoramic image generation are provided, with representative artifacts highlighted in red boxes.
    }
    \label{fig:ablations_full_comparison}
\end{figure*}

\paragraph{Implementation Details.}
We implement \textit{Canvas360} on top of FLUX.1-dev~\citep{flux}.
We adopt parameter-efficient fine-tuning by injecting LoRA~\citep{lora} into the attention blocks and the in/out embedding layers, using rank $r{=}64$, scaling factor $\alpha{=}64$, and LoRA dropout of $0.10$.
All models are fine-tuned with FP16 mixed precision on 8 NVIDIA H20 GPUs.
We optimize only the LoRA-injected trainable parameters using AdamW~\citep{adamw} with a learning rate of $1\times10^{-5}$, $\beta_{1}=0.9$, $\beta{2}=0.999$, $\epsilon=10^{-6}$, and weight decay $0$.
Training runs for 25 epochs with a per-GPU batch size of 1 and gradient accumulation of 3 (effective batch size $=24$).
We use a constant learning-rate schedule with a 10\% warmup (by steps) and set the training guidance scale to $1.0$.
The main experiments are conducted at a resolution of $1024{\times}2048$, while the ablation studies are performed at $512{\times}1024$.
For inference, we use classifier-free guidance with scale 3.0 and 28 sampling steps.

\paragraph{Evaluation Metrics.}
Following prior work, we evaluate our method with a diverse set of complementary metrics.
We measure realism using Fréchet Inception Distance (FID)~\citep{fid} and its variants, FID\textsubscript{pole} and FID\textsubscript{equ} (following SMGD~\citep{smgd}), to evaluate polar distortion and equatorial perspective quality.
Since FID uses an Inception model trained on perspective images and may under-reflect panoramic properties, we additionally report Fréchet Auto-Encoder Distance (FAED)~\citep{faed}, which is tailored for panoramas.
For diversity, we use Inception Score (IS)~\citep{is} and replace the standard Inception-v3~\citep{inceptionv3} with a Places365-pretrained ResNet~\citep{resnet,places365} to better match our scene-centric data.
We measure text--image alignment with CLIP Score (CS)~\citep{clip}, and report Q-Align (QA)~\citep{q_align}, BRISQUE~\citep{brisque}, and NIQE~\citep{niqe} for perceptual quality, following HunyuanWorld~\citep{hunyuanworld}.


\section{Full Comparison}
\label{appendix:full_comparison}

\begin{table*}[ht]
\centering
\caption{
Quantitative comparison of left-to-right boundary consistency. 
Following PanoFormer~\citep{panoformer}, we extend LRCE to RGB panoramas by measuring the discrepancy between the left and right boundary regions. 
Lower LRCE-RGB indicates better boundary consistency.
}
\label{tab:appendix_lrce_rgb}
\begingroup
\small
\setlength{\tabcolsep}{4pt}
\renewcommand{\arraystretch}{1.12}
\begin{tabular}{lcccccccc}
\toprule
Metric 
& PanFusion 
& SMGD 
& PAR 
& WorldGen 
& LayerPano3D 
& HunyuanWorld 
& DiT360 
& Ours \\
\midrule
LRCE-RGB$\downarrow$ 
& 0.0154 
& 0.0146 
& 0.0171 
& 0.0152 
& 0.0188 
& 0.0094 
& 0.0101 
& \textbf{0.0063} \\
\bottomrule
\end{tabular}
\endgroup
\end{table*}

We include the full qualitative comparisons in~\cref{fig:ablations_full_comparison}, where typical failure patterns are marked with red boxes. 
Prior panorama generators based on structural heuristics or autoregressive formulations often struggle to simultaneously maintain sharp details and clean global structure, leading to noisy textures and local distortions. 
Recent DiT-based methods improve overall fidelity, but fine-grained content remains fragile under ERP distortions, and discontinuities near the seam are still common. 
DiT360~\citep{feng2025dit360} mitigates some of these issues with cube-space supervision; nevertheless, projecting between cube and ERP introduces information loss, particularly for latent features, resulting in residual artifacts and imperfect long-range consistency. 
Across diverse prompts, our approach yields more panorama-consistent generations, exhibiting sharper distortion-aware details and markedly improved seam continuity.

To further assess boundary consistency, we additionally evaluate the left-to-right consistency of generated RGB panoramas. 
Following PanoFormer~\citep{panoformer}, we extend LRCE to RGB panoramas by measuring the discrepancy between the left and right boundary regions. 
As reported in~\cref{tab:appendix_lrce_rgb}, our method achieves the lowest LRCE-RGB among all compared methods, demonstrating the best left-to-right consistency. 
This result quantitatively supports the qualitative observations in~\cref{fig:ablations_full_comparison}, showing that our method better preserves seamless horizontal continuity and reduces boundary artifacts in ERP panoramic generation.

\section{Ablations}
\label{appendix:ablation}

\begin{figure}[ht]
    \centering
    \includegraphics[width=\linewidth]{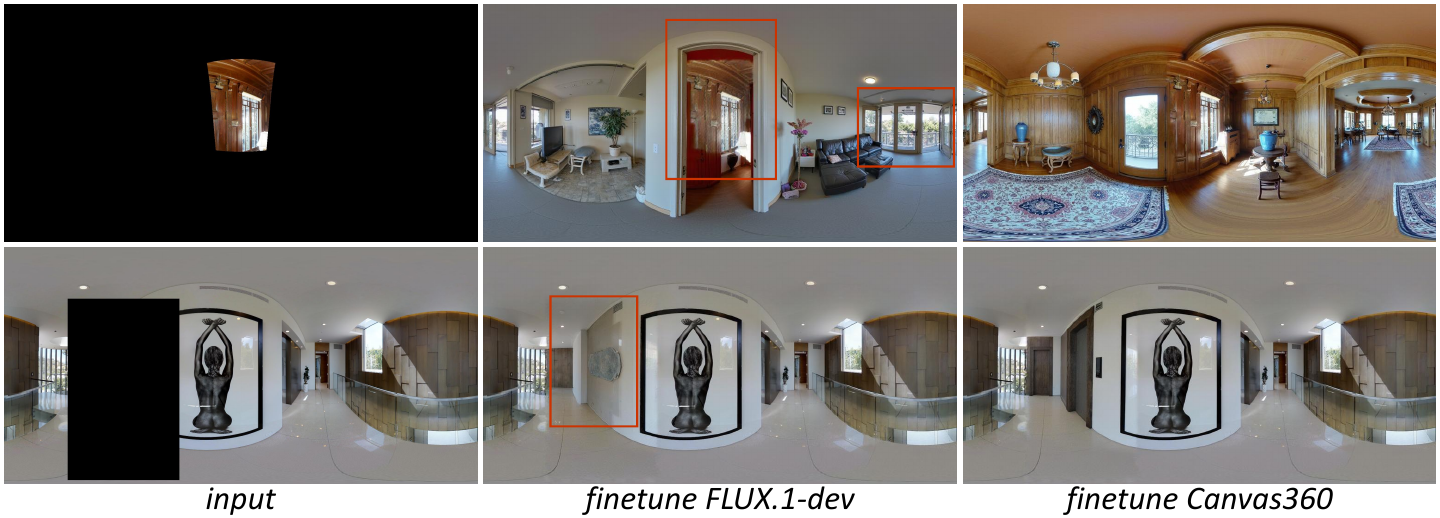}
    \caption{
    Qualitative ablations for model backbone.
    }
    \label{fig:ablation_for_baseline}
\end{figure}

\begin{figure*}[ht]
    \centering
    \includegraphics[width=\linewidth]{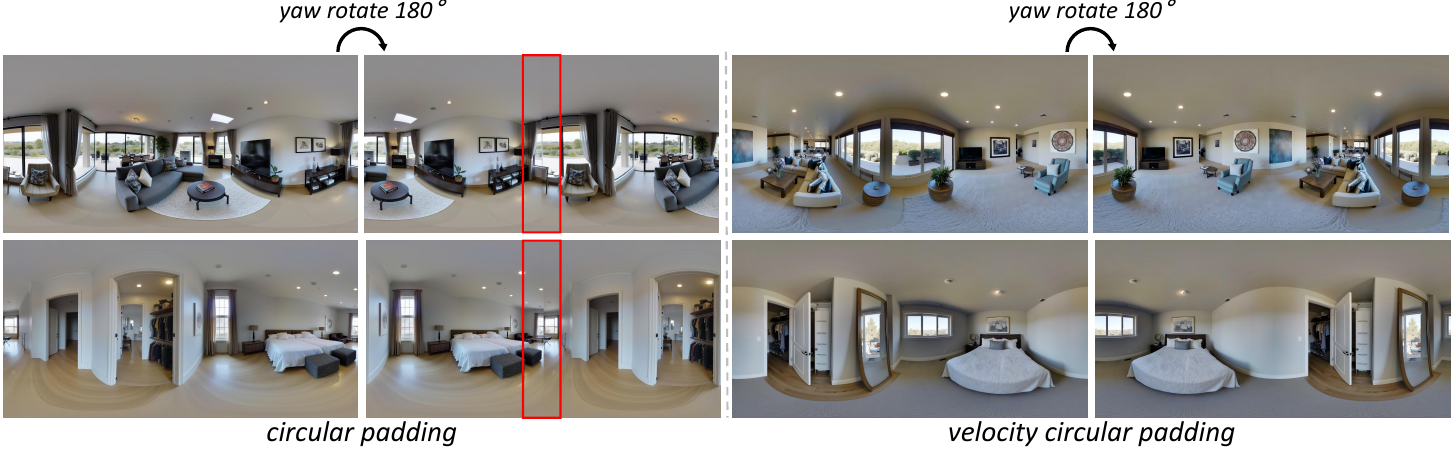}
    \caption{
    Qualitative ablations for velocity circular padding.
    }
    \label{fig:ablations_for_vcp}
\end{figure*}

\begin{figure*}[ht]
    \centering
    \includegraphics[width=\linewidth]{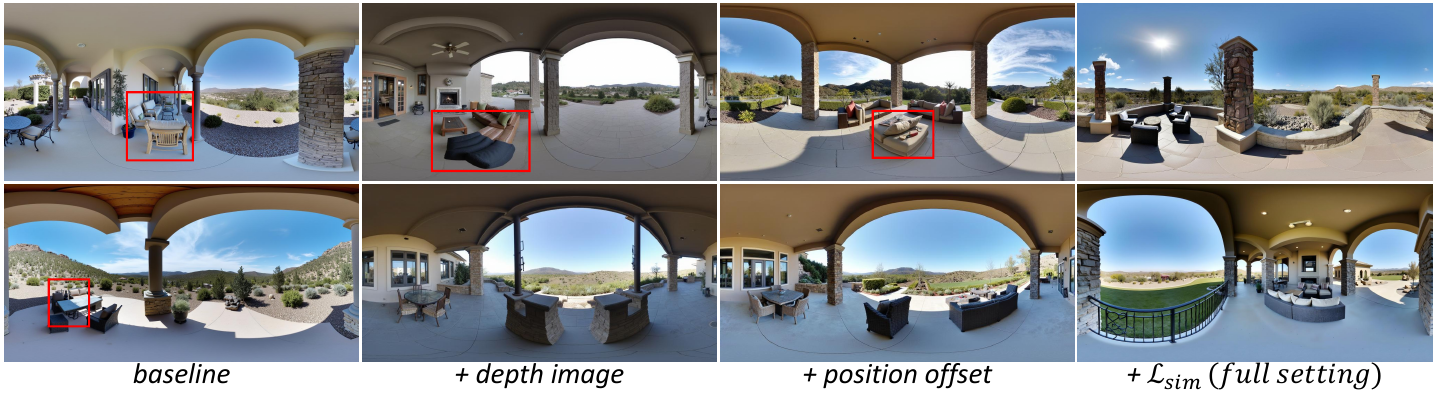}
    \caption{
    Qualitative ablations for parallel depth generation.
    }
    \label{fig:ablations_for_pdg}
\end{figure*}

\begin{table*}[ht]
\centering
\caption{
Complete quantitative ablation study. 
We report FID, FID$_{pole}$, FID$_{equ}$, FAED, IS, CS, QA$_{qua}$, QA$_{ae}$, BRISQUE, and NIQE. 
Here, B, VCP, D, and PO denote the baseline, velocity circular padding, depth image, and position offset, respectively. 
All variants are built upon the same baseline. VCP is evaluated independently, while the depth branch is cumulative: PO is added on top of D, and $\mathcal{L}_{\text{sim}}$ is further added on top of both.
Red and orange cells indicate the column-wise top-1 and top-2 results, respectively, according to the optimization direction of each metric.
Lower values are better for FID, FID$_{pole}$, FID$_{equ}$, FAED, BRISQUE, and NIQE, while higher values are better for IS, CS, QA$_{qua}$, and QA$_{ae}$.
}
\label{tab:appendix_full_ablation}
\begingroup
\scriptsize
\setlength{\tabcolsep}{5pt}
\newcommand{\best}[1]{\cellcolor{red!15}#1}
\newcommand{\second}[1]{\cellcolor{orange!15}#1}
\begin{tabular}{@{}lcccccccccc@{}}
\toprule
Methods 
& FID$\downarrow$ 
& FID$_{pole}\downarrow$
& FID$_{equ}\downarrow$
& FAED$\downarrow$ 
& IS$\uparrow$ 
& CS$\uparrow$ 
& QA$_{qua}\uparrow$
& QA$_{ae}\uparrow$
& BRISQUE$\downarrow$
& NIQE$\downarrow$ \\
\midrule
B
& \best{51.16} & \best{53.33} & \best{28.10} & 5.37 & 1.83 & 34.62 & 4.40 & 3.97 & \second{14.62} & 3.86 \\
B + VCP
& 53.53 & \second{54.13} & \second{28.49} & 5.43 & 1.77 & 34.58 & \second{4.52} & 3.92 & \best{13.85} & \best{3.79} \\
B + D
& 51.57 & 55.47 & 29.02 & \best{4.74} & \best{1.98} & \best{34.81} & 4.23 & 3.41 & 17.84 & 3.85 \\
B + D + PO
& 57.04 & 62.82 & 29.44 & \second{4.81} & 1.85 & 34.72 & 4.44 & \second{4.13} & 23.64 & 4.12 \\
B + D + PO + $\mathcal{L}_{\text{sim}}$
& \second{51.48} & 55.63 & 29.74 & 4.93 & \second{1.88} & \second{34.73} & \best{4.71} & \best{4.20} & 16.23 & \second{3.84} \\
\bottomrule
\end{tabular}
\endgroup
\end{table*}

We provide more detailed ablation results in this appendix, including qualitative studies on velocity circular padding and parallel RGB--depth generation, as well as a complete quantitative ablation in~\cref{tab:appendix_full_ablation}.

\paragraph{Velocity Circular Padding.}
\cref{fig:ablations_for_vcp} shows the qualitative ablation on velocity circular padding. 
Compared with naive circular padding, the proposed strategy synchronizes ghost-column
features with their circular counterparts while assigning them continuous
longitude indices. 
This exposes the horizontal wrap-around boundary as local
coordinate transitions, leading to more seamless panorama boundaries.
The quantitative results in ~\cref{tab:appendix_full_ablation} provide a more nuanced observation. 
Adding velocity circular padding improves the no-reference image quality metrics,
i.e., BRISQUE and NIQE, and also increases QA$_{\mathrm{qua}}$, indicating
better perceptual quality and fewer local seam artifacts. 
However, it does not
necessarily improve distribution-level metrics such as FID, FID$_{\mathrm{pole}}$,
FID$_{\mathrm{equ}}$, and FAED. 
This is expected because velocity circular
padding mainly targets local boundary continuity rather than global distribution
matching or semantic fidelity.


\noindent\textbf{Parallel RGB--Depth Generation.}
\cref{fig:ablations_for_pdg} presents the ablation study on parallel RGB--depth generation.
Depth supervision, positional offsets, and similarity regularization are designed as a coupled training strategy, where depth provides geometric cues and the other two components stabilize cross-modal optimization.
As shown in~\cref{tab:appendix_full_ablation}, introducing depth supervision clearly improves the panorama-oriented FAED metric, reducing it from 5.37 to 4.74.
Since FAED can be regarded as a panorama-aware variant of FID that better captures the feature distribution of omnidirectional images, this improvement indicates that auxiliary depth supervision provides useful geometric guidance for panoramic generation.
However, depth supervision alone is not sufficiently stable in our setting.
Although it improves FAED, IS, CS, and NIQE, it degrades QA$_{\text{qua}}$, QA$_{\text{ae}}$, and BRISQUE, and we empirically observe that depth-only training can produce over-darkened regions, near-black failure cases, and visible artifacts under certain configurations.
This instability is likely caused by the noise and distortion in ERP depth signals, which makes direct RGB--depth co-training prone to modality entanglement and unstable optimization.

To mitigate this issue, we introduce positional offsets to better separate RGB and depth tokens in the positional encoding space.
This design improves QA$_{\text{ae}}$ from 3.41 to 4.13, suggesting that separating the two modalities helps recover more favorable visual and aesthetic properties.
Nevertheless, positional offsets alone do not fully resolve the optimization instability, as reflected by the degraded FID-family and no-reference quality metrics.
We therefore further introduce the similarity regularization $\mathcal{L}_{\text{sim}}$, which stabilizes RGB--depth co-training by preventing the RGB and depth branches from becoming overly coupled or collapsing into similar representations.
With $\mathcal{L}_{\text{sim}}$, the model achieves the best QA$_{\text{qua}}$ and QA$_{\text{ae}}$ scores within this RGB--depth ablation, while maintaining competitive FAED, IS, CS, and NIQE performance.
These results suggest that depth supervision contributes useful panorama-aware geometric cues, while positional offsets and similarity regularization are important for improving training robustness, suppressing degenerate dark-output cases, and producing visually more reliable panoramic generations.

\noindent\textbf{Backbone Design.}
\cref{fig:ablation_for_baseline} provides evidence that \textit{Canvas360} learns stronger panoramic priors that transfer to image completion. 
We fine-tune FLUX.1-dev~\cite{flux} and \textit{Canvas360}, respectively, under the same completion setting. 
FLUX.1-dev tends to introduce blur and artifacts in inpainting, and its outpainting results are less consistent with the conditioning signal, often exhibiting perspective-biased patterns. 
In contrast, \textit{Canvas360} produces cleaner and more panorama-consistent completions, indicating that depth-augmented pretraining helps the model internalize geometry-aware panoramic distortions.

\section{Full Comparisons on In-context Panoramic Generation}
\label{appendix:full_comparison_on_in-context_panoramic_generation}

We conducted additional experiments for in-context panoramic generation. 
For each experiment, we follow the corresponding experimental setting and generate 500 results for evaluation.

\paragraph{Style Transfer.}

\begin{figure*}[ht]
    \centering
    \includegraphics[width=1\linewidth]{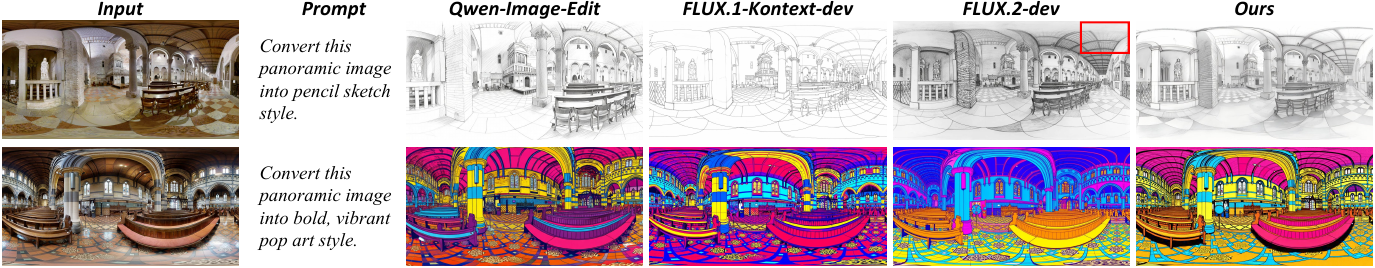}
    \caption{
    Qualitative comparisons for style transfer.
    }
    \label{fig:app_q_f_style_trans}
\end{figure*}

\begin{table}[ht]
\centering
\caption{Quantitative comparison for style transfer. We report CP (Content Preservation), SR (Style Resemblance), and OV (Overall Vision). Higher values indicate better performance.}
\label{tab:appendix_qc_f_style_trans}
\begin{tabular}{lccc}
\toprule
Method & CP$\uparrow$ & SR$\uparrow$ & OV$\uparrow$ \\
\midrule
FLUX.1-Kontext-dev & 0.457 & 0.482 & 0.467 \\
FLUX.2-dev         & 0.490 & \textbf{0.503} & 0.495 \\
Qwen-Image-Edit    & 0.464 & 0.483 & 0.471 \\
Ours               & \textbf{0.502} & 0.491 & \textbf{0.497} \\
\bottomrule
\end{tabular}
\end{table}

We compare our approach with three representative image editing baselines, including FLUX.1-Kontext-dev~\citep{flux1kontext}, FLUX.2-dev~\citep{flux2}, and Qwen-Image-Edit~\cite{qwen_img}. 
Following SRQE~\citep{srqe}, we adopt three evaluation metrics: CP (Content Preservation), SR (Style Resemblance), and OV (Overall Vision). 
CP measures whether the generated panorama preserves the content and structural layout of the input image, SR evaluates the resemblance between the generated result and the target style, and OV reflects the overall visual quality and consistency.
The qualitative results are shown in~\cref{fig:app_q_f_style_trans}, and the quantitative results are reported in~\cref{tab:appendix_qc_f_style_trans}.

The qualitative results show that our method better preserves the geometric structure and content layout of the input panoramas, while producing visually coherent stylized results. 
The quantitative results further support this finding. 
Our method achieves the best performance in CP and OV, demonstrating stronger content preservation and overall visual quality. 
Although FLUX.2-dev obtains a slightly higher SR score, our method still achieves competitive style resemblance, indicating a better balance between faithful panoramic structure preservation and effective style transfer.

\paragraph{Inpainting and Outpainting.}

\begin{table}[ht]
\centering
\caption{Quantitative comparison for inpainting and outpainting. We report LPIPS, FAED, and PSNR. Lower LPIPS and FAED indicate better performance, while higher PSNR indicates better performance.}
\label{tab:appendix_qc_f_inpaint_outpaint}
\begin{tabular}{llccc}
\toprule
Task & Method & LPIPS$\downarrow$ & FAED$\downarrow$ & PSNR$\uparrow$ \\
\midrule
\multirow{4}{*}{Inpainting}
& Flux.1-Fill-dev & 0.171 & 0.461 & 24.23 \\
& PAR              & 0.158 & 0.455 & 23.76 \\
& PanoDiffusion    & 0.147 & 0.523 & 23.93 \\
& Ours             & \textbf{0.096} & \textbf{0.371} & \textbf{25.87} \\
\midrule
\multirow{4}{*}{Outpainting}
& Flux.1-Fill-dev & 0.509 & 1.916 & 16.32 \\
& PAR              & 0.553 & 1.849 & 16.71 \\
& PanoDiffusion    & 0.674 & 1.989 & 15.21 \\
& Ours             & \textbf{0.416} & \textbf{1.791} & \textbf{17.16} \\
\bottomrule
\end{tabular}
\end{table}

The baselines for inpainting and outpainting are the same as those in the main paper. 
We evaluate the generated panoramas using LPIPS~\citep{lpips}, FAED~\citep{faed}, and PSNR. 
LPIPS measures perceptual similarity, FAED evaluates the distribution-level fidelity of generated panoramic images, and PSNR reflects pixel-level reconstruction quality.
The quantitative results are reported in~\cref{tab:appendix_qc_f_inpaint_outpaint}.
The results show that our method consistently outperforms all baselines on both inpainting and outpainting tasks. 
Across both inpainting and outpainting, our method achieves the best results on all three metrics, yielding lower LPIPS and FAED as well as higher PSNR than all baselines. 
These consistent improvements indicate that our method can better preserve perceptual quality and distributional fidelity while producing more accurate reconstructions, demonstrating its effectiveness for both completing missing regions and extending panoramic content with coherent structure and visual consistency.

\paragraph{Editing.}

\begin{figure*}[ht]
    \centering
    \includegraphics[width=1\linewidth]{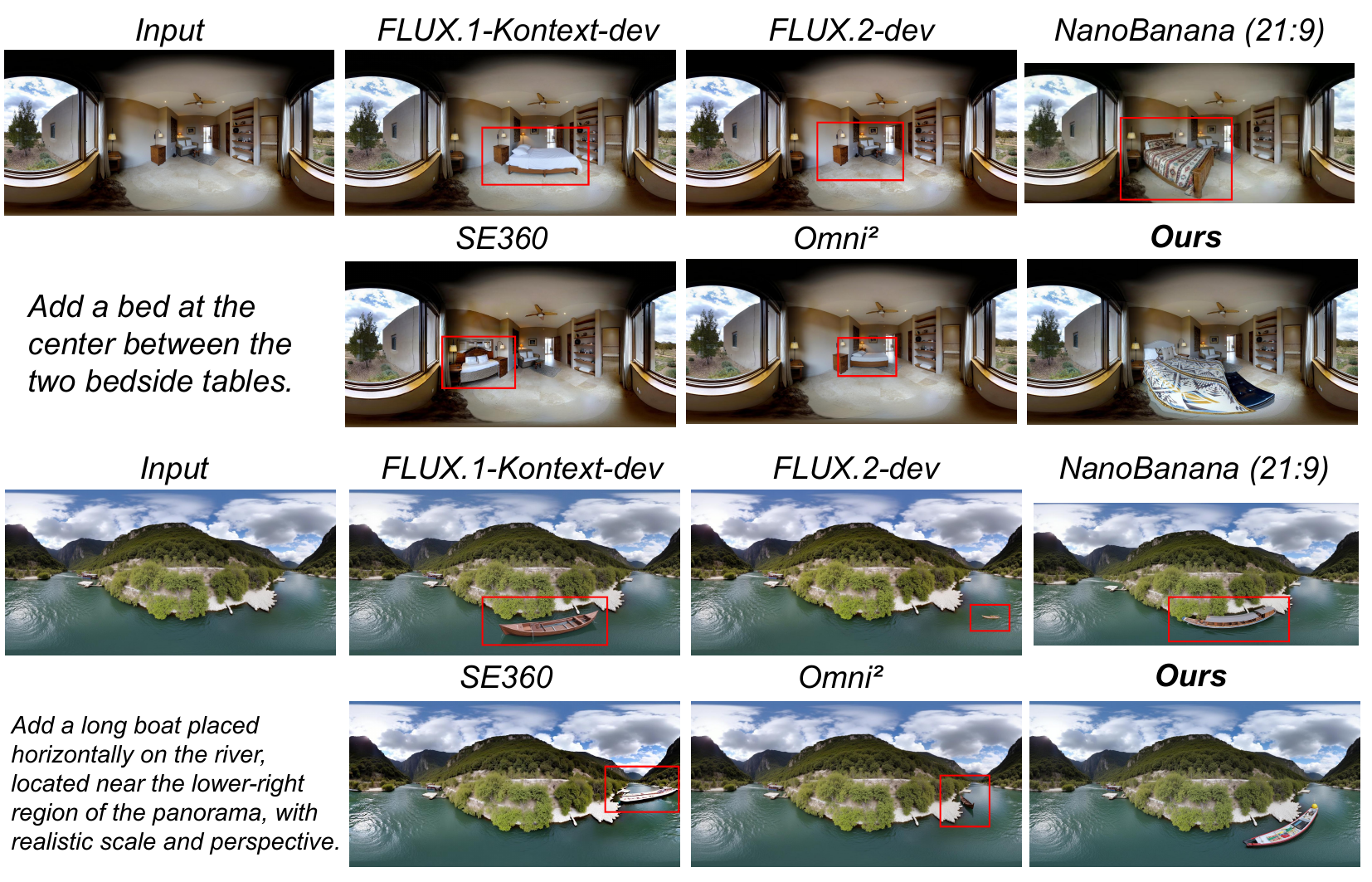}
    \caption{
    Qualitative comparisons for editing.
    }
    \label{fig:app_q_f_editing}
\end{figure*}

\begin{table}[ht]
\centering
\caption{Quantitative comparison for editing. We report LPIPS, FAED, and PSNR. Lower LPIPS and FAED indicate better performance, while higher PSNR indicates better performance.}
\label{tab:appendix_qc_f_editing}
\begin{tabular}{lccc}
\toprule
Method & LPIPS$\downarrow$ & FAED$\downarrow$ & PSNR$\uparrow$ \\
\midrule
FLUX.1-Kontext-dev & 0.102 & 0.458 & 25.77 \\
FLUX.2-dev         & 0.099 & 0.410 & 26.17 \\
NanoBanana         & 0.094 & 0.395 & 25.91 \\
SE360              & 0.138 & 0.386 & 25.16 \\
Omni2              & 0.105 & 0.392 & 25.03 \\
Ours               & \textbf{0.084} & \textbf{0.358} & \textbf{26.40} \\
\bottomrule
\end{tabular}
\end{table}

For editing, we compare our method with FLUX.1-Kontext-dev~\citep{flux1kontext}, FLUX.2-dev~\citep{flux2}, NanoBanana~\citep{banana}, and additionally include two panoramic editing baselines, SE360~\citep{zhong2025se360} and Omni2~\citep{omni2}. 
We evaluate the generated panoramas using LPIPS, FAED~\citep{faed}, and PSNR. 
The qualitative results are shown in~\cref{fig:app_q_f_editing}, and the quantitative results are reported in~\cref{tab:appendix_qc_f_editing}.

The qualitative results show that our method produces more faithful editing results while better preserving the panoramic geometry and surrounding content consistency. 
The quantitative results further support this finding. 
Our method achieves the best performance across all three metrics, with the lowest LPIPS and FAED as well as the highest PSNR. 
These improvements demonstrate that our method can perform effective panoramic editing while maintaining stronger perceptual quality, distributional fidelity, and reconstruction accuracy.

\section{More Results}
\label{appendix:more_app}

\begin{figure*}[ht]
    \centering
    \includegraphics[width=1\linewidth]{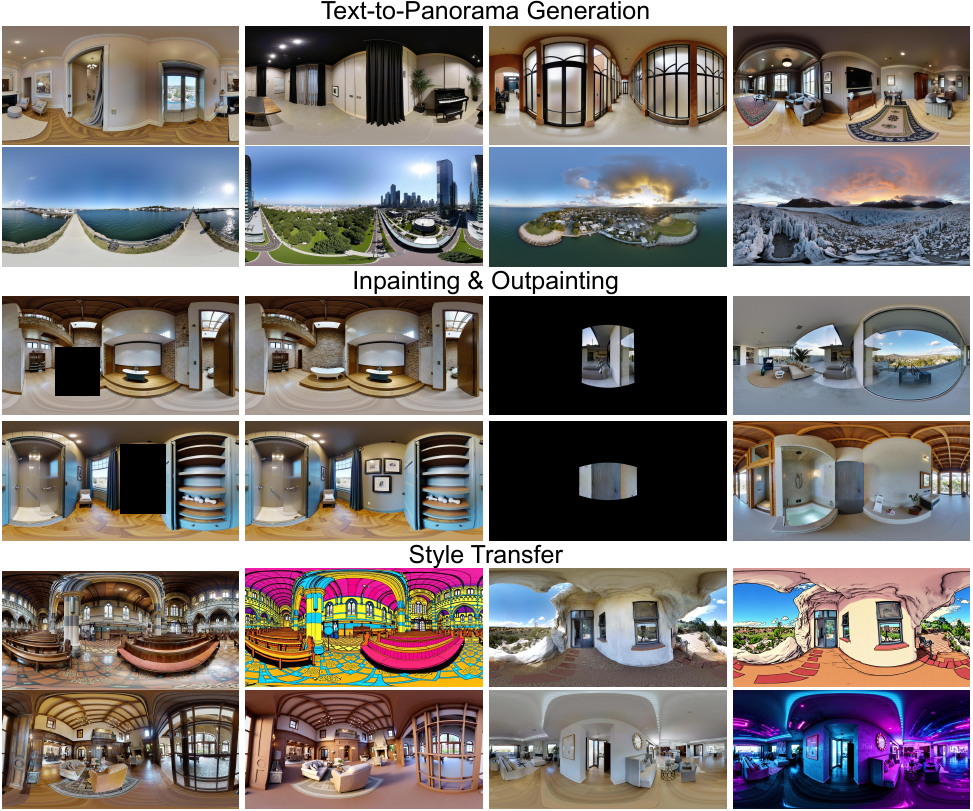}
    \caption{
    More results of \textit{Canvas360}.
    }
    \label{fig:app_more_results}
\end{figure*}

\begin{figure*}[ht]
    \centering
    \includegraphics[width=1\linewidth]{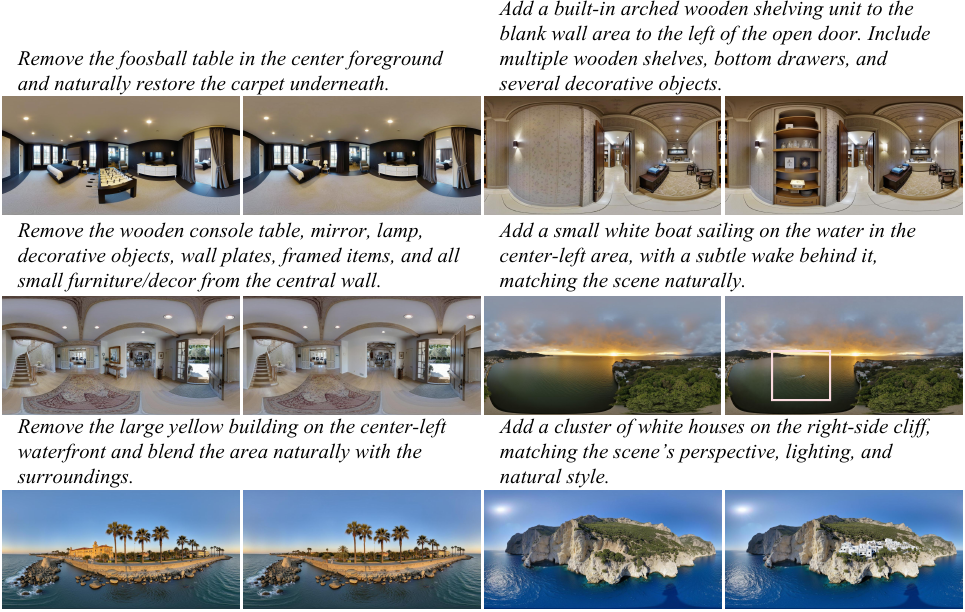}
    \caption{
    More results of \textit{Canvas360}.
    }
    \label{fig:app_more_editing_results}
\end{figure*}

We provide additional results on in-context panoramic generation in \cref{fig:app_more_results,fig:app_more_editing_results}.
Across all tasks, \textit{Canvas360} consistently produces high-fidelity and visually coherent panoramas, with distortion-consistent details and strong seam continuity. 
These results further highlight \textit{Canvas360}'s robust panorama-aware generation capability and validate that our framework learns geometry-consistent panoramic priors that generalize across diverse in-context scenarios.

\section{Limitations and Future Work}
\label{appendix:limit}

Despite the strong performance of \textit{Canvas360}, our approach still has limitations. 
First, our training corpus remains imbalanced across scene types, and high-quality panoramic data for certain categories is relatively scarce. 
As a result, \textit{Canvas360} can underperform on underrepresented cases such as high-resolution human faces and text-rich signage, particularly in heavily distorted ERP regions. 
In future work, we will expand and rebalance the dataset to strengthen the panoramic prior in challenging categories and improve the model's robustness to rare scene contents and severe geometric distortions.

\section{Broader impacts}
\label{appendix:broader_impacts}

This work can support benign applications such as VR/AR authoring, simulation, digital-twin prototyping, immersive scene creation, and 360-degree content design by improving the geometric consistency and visual fidelity of panoramic generation. 
However, as with other generative vision models, it may also be misused for deceptive scene manipulation, synthetic visual misinformation, or unauthorized content creation, and large-scale data curation may involve licensing or attribution concerns. 
Although our method focuses on panoramic scenes rather than identity-centric or face-oriented generation, responsible use remains important. We encourage safeguards such as data filtering, provenance tracking, watermarking, transparent attribution, and human review before deployment in sensitive settings.

\section{Safeguards for Responsible Release}
\label{appendix:safeguards}

Our work involves panoramic image generation and large-scale data curation, which may carry potential misuse risks similar to other generative vision models. To support responsible release, we apply data filtering during dataset construction to remove unsafe, sensitive, or low-quality samples, and we focus on panoramic scene-level content rather than identity-centric or face-oriented imagery. This design reduces risks related to privacy, impersonation, and personal attribute generation.

For released resources, we will provide usage guidelines that discourage deceptive scene manipulation, synthetic visual misinformation, and unauthorized content generation. We also encourage downstream users to adopt safeguards such as provenance tracking, watermarking, transparent attribution, and human review, especially before deploying the model or generated content in sensitive applications. Since the full-scale dataset is large, we will release it with accompanying documentation describing data sources, preparation procedures, filtering steps, and intended-use restrictions.

\section{Human Preference Study Details}
\label{appendix:user_study}

We provide additional details of the human preference study used in~\cref{tab:user_study}. 
The study was designed to evaluate perceptual preferences among different panoramic generation methods. 
For each question, participants were shown four anonymized candidate panoramic images generated by different methods and were asked to select the image that best satisfied the displayed evaluation criterion.

\paragraph{Evaluation criteria.}
Participants evaluated the generated results under four criteria: text alignment (TA), boundary continuity (BC), panorama awareness (PA), and overall quality (OQ). 
Text alignment measures whether the generated panorama is consistent with the input prompt. 
Boundary continuity measures whether the panorama is seamless near the horizontal wrap-around boundary. 
Panorama awareness measures whether the image properly reflects panoramic geometry, including spherical distortion and wide-field spatial layout. 
Overall quality measures the general perceptual fidelity, realism, and visual appeal of the generated panorama.

\begin{figure}[t]
    \centering
    \includegraphics[width=0.95\linewidth]{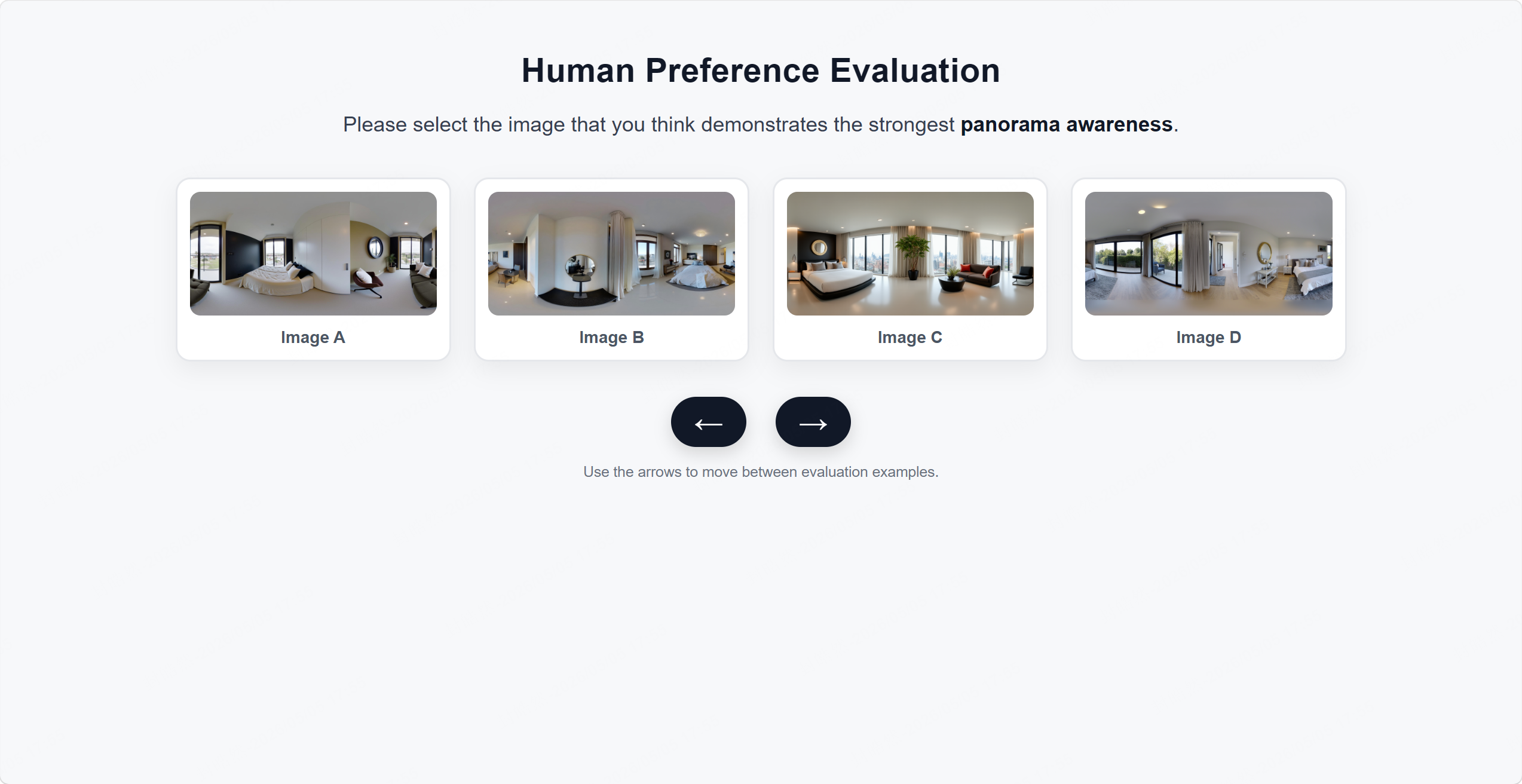}
    \caption{
    Screenshot of the human preference study interface. 
    For each question, participants were shown four anonymized candidate panoramic images and selected the one that best satisfied the displayed evaluation criterion.
    }
    \label{fig:user_study_interface}
\end{figure}

\paragraph{Study interface.}
Fig.~\ref{fig:user_study_interface} shows the interface used in our human preference study. 
Each page presented one evaluation criterion and four anonymized candidate images. 
Participants selected one image from the four candidates according to the displayed criterion. 
The method names were hidden during evaluation to reduce potential bias.

\paragraph{Participants and procedure.}
The study was conducted with internal volunteer participants from the authors' organization. 
Participants were informed of the study procedure before starting the evaluation. 
Participation was voluntary, and participation or non-participation had no effect on employment, compensation, or performance evaluation. 
Each participant compared generated panoramic images through four-choice questions. 
The collected responses were used only for aggregate statistical analysis.

\paragraph{Risk, privacy, and review.}
The study involved minimal risk because participants only compared generated panoramic images and did not interact with sensitive content or provide personal information beyond image preference choices. 
We did not collect personally identifiable information, sensitive attributes, private data, or free-form personal responses. 
All results were aggregated across participants before analysis and reporting. 
The study was reported and reviewed through the authors' organizational review process. 
To preserve anonymity in the initial submission, institution-identifying details are omitted.


\end{document}